%% file: main.tex
\documentclass[times, review, 10pt]{elsarticle}

\usepackage{amssymb}
\usepackage{amsmath}

\DeclareMathOperator*{\argmin}{arg\,min}

\usepackage{dsfont}

\usepackage{algorithm}
\usepackage{algpseudocode}
\usepackage{xcolor}
\usepackage{booktabs} 
\usepackage{multirow}
\usepackage{float}
\usepackage{placeins}
\usepackage{graphicx}
\usepackage{subcaption}

\usepackage{hyperref}

\journal{Pattern Recognition}

\begin{document}

\begin{frontmatter}

\title{Conformalized High-Density Quantile Regression via Dynamic Prototypes-based Probability Density Estimation}

% use optional labels to link authors explicitly to addresses:
\author[label1]{Batuhan Cengiz}
\author[label2]{Halil Faruk Karagoz}
\author[label3]{Tufan Kumbasar}
    
\affiliation[label1]{organization={
            AI and Data Engineering Department,
            Istanbul Technical University},
            city={Istanbul},
            country={Türkiye}}

\affiliation[label2]{organization={
            Computer Engineering Department,
            Istanbul Technical University},
            city={Istanbul},
            country={Türkiye}}

\affiliation[label3]{organization={
            Artificial Intelligence and Intelligent Systems Laboratory,
            Istanbul Technical University},
            city={Istanbul},
            country={Türkiye}}
            
% \author{Author1, Author2, Author3...} %% Author name

% %% Author affiliation
% \affiliation{organization={OrganizationA},%Department and Organization
%             addressline={}, 
%             city={},
%             postcode={}, 
%             state={},
%             country={}}

%% Abstract
\begin{abstract}
% %% Text of abstract

Recent methods in quantile regression have adopted a classification perspective to handle challenges posed by heteroscedastic, multimodal, or skewed data by quantizing outputs into fixed bins. Although these regression-as-classification frameworks can capture high-density prediction regions and bypass convex quantile constraints, they are restricted by quantization errors and the curse of dimensionality due to a constant number of bins per dimension. To address these limitations, we introduce a conformalized high-density quantile regression approach with a dynamically adaptive set of prototypes. Our method optimizes the set of prototypes by adaptively adding, deleting, and relocating quantization bins throughout the training process. Moreover, our conformal scheme provides valid coverage guarantees, focusing on regions with the highest probability density. Experiments across diverse datasets and dimensionalities confirm that our method consistently achieves high-quality prediction regions with enhanced coverage and robustness, all while utilizing fewer prototypes and memory, ensuring scalability to higher dimensions. 

\noindent
The code is available at \href{https://github.com/batuceng/max_quantile}{\color{magenta}https://github.com/batuceng/max\_quantile} .
\end{abstract}

% %%Graphical abstract
% \begin{graphicalabstract}
% %\includegraphics{grabs}
% \end{graphicalabstract}

%%Research highlights
%% Highlights page is commented for ArXiv
% \begin{highlights}
% \item Scalable regression-as-classification model for high-density, multi-target data
% \item Dynamic prototype regions enhance precision in complex, multimodal data
% \item Conformalized approach with adaptive coverage for high-confidence predictions
% \end{highlights}

%% Keywords
\begin{keyword}
%% keywords here, in the form: keyword \sep keyword
conformal prediction \sep prediction regions \sep multi-target \sep prototypes  
%% PACS codes here, in the form: \PACS code \sep code

%% MSC codes here, in the form: \MSC code \sep code
%% or \MSC[2008] code \sep code (2000 is the default)

\end{keyword}

\end{frontmatter}

%% Add \usepackage{lineno} before \begin{document} and uncomment 
%% following line to enable line numbers
%% \linenumbers

%% main text
%%

%% Use \section commands to start a section
\section{Introduction}
\label{sec_intro}

% ::Flow of Intro::
% Start by quantile regression
% CP procedure and CQR
% Regression-as-classification / High Density Region (Non-Convex Quantile)
% Curse of dimensionality (Grid spaced bins does not scale)

Regression analysis is a fundamental tool for modeling relationships between variables, often predicting a central tendency such as the mean. However, in many real-world applications, understanding extreme values or the range of possible outcomes is crucial, especially in safety-critical contexts where risks must be tightly managed. Quantile regression \cite{Koenker1978regression} extends traditional regression by estimating conditional quantiles, allowing for the prediction of boundaries or worst-case scenarios rather than just average trends. This approach is invaluable for safety-critical applications, where anticipating extremes—such as peak loads in power grids \cite{Liu2017probabilistic}, maximum delays in air traffic \cite{dalmau2024probabilistic}, or critical health indicators in medical treatment\cite{austin2005use}—can prevent failures and protect lives. By capturing the variability in data distributions, quantile regression provides predictions that support more resilient decision-making across fields like autonomous driving, healthcare\cite{austin2005use, lu2022improving}, finance\cite{arias2002individual}, and industrial automation \cite{yin2024reliable}.

% Introduce Conformal Prediction, Tell about conformalized quantile regression, uncertainity estimation

% Tell disadvantage of convex quantiles for bimodal data etc. it may not always have the regions with most likelihood, it expands over median gibi bir şey. Regression-as-classification R2CCP

% Problem with fixed bins, unscalable to higher dimension. Introduce our method
While quantile regression can provide predictive intervals, it does not guarantee the accuracy of these intervals. This challenge is not unique to quantile regression; it also affects traditional regression and classification methods. Conformal prediction (CP) \cite{vovk1999machine, vovk2005algorithmic, angelopoulos2021gentle} addresses this issue by offering reliable uncertainty estimates through prediction regions that adapt to the data without relying on strict distributional assumptions. Building on this, Conformalized Quantile Regression (CQR) \cite{romano2019conformalized} combines conformal prediction with quantile regression to create prediction regions that cover true outcomes at a specified confidence level, regardless of the data distribution.

%This framework is particularly useful in high-dimensional and complex applications, as it allows for the direct estimation of high-density regions where extreme values are likely to occur. 

% Although conformalized quantile regression (CQR) provides coverage guarantees, it has several limitations, such as generating convex intervals, creating overly large intervals, and struggling to handle bimodal or multimodal data effectively. To address these issues, recent studies have proposed alternative approaches that better adapt to complex data distributions and provide more efficient prediction regions. For example \cite{izbicki2022cd}

Although Conformalized Quantile Regression (CQR) provides coverage guarantees through conformal prediction, it faces notable limitations, one of which is the convexity constraint. By definition, quantiles form convex intervals centered around points like the median, which can be restrictive for various data modalities, particularly multimodal distributions \cite{lei2014distribution}. In these scenarios, quantile bands may expand excessively to meet the required coverage, with this growth becoming exponential in multivariate settings. To address this, Kong et al. \cite{kong2012quantile} proposed joint boundaries for tighter quantiles, yet this solution remains constrained by convexity. More recently, Feldman et al. \cite{feldman2023calibrated} proposed a conditional variational autoencoder (CVAE) \cite{sohn2015learning} to predict quantile intervals within a latent space, allowing for non-convex quantile regions through the decoder, which can better accommodate complex data distributions.

The convexity constraint can also limit the adaptability of quantile regression to complex data distributions, especially in multimodal or high-dimensional cases. To address this, Izbicki et al. \cite{izbicki2022cd} proposed an approach using a union of split quantiles, enabling non-convex prediction regions. The same work also explored using High-Density Regions (HDRs) as an alternative representation of quantiles. Defined as non-convex regions, HDRs \cite{hyndman1995highest,Hyndman1996computing} cover a $(1-\alpha)$ probability subset of the space, ensuring that no points outside the region have a higher density than those within. In a different line of work, Guha et al. \cite{guha2024conformal} introduced the Regression-as-Classification (RasC) framework to estimate non-convex quantiles by reframing the quantile regression task as a classification problem, thus enhancing the flexibility of non-convex quantile estimation. This approach allows them to conformalize their method using conformal prediction (CP) algorithms for classification \cite{romano2020classification, angelopoulos2020uncertainty}.

The RasC approach by Guha et al. \cite{guha2024conformal}, while effective, has several limitations. It relies on a fixed number of quantization bins to convert regression into classification, and its interpolation from discrete probability mass functions to continuous probability density functions is feasible only for single-target quantile regression. Even without considering integral interpolation, fixed bins or grid-spaced quantiles tend to scale exponentially in computational cost and lack the flexibility required to adapt to intricate data structures. This highlights the need for innovative methods that can efficiently scale and adapt to capture complex quantile structures, particularly in high-dimensional spaces.

\begin{figure}[t]
    \centering
    \includegraphics[width=0.99\textwidth]{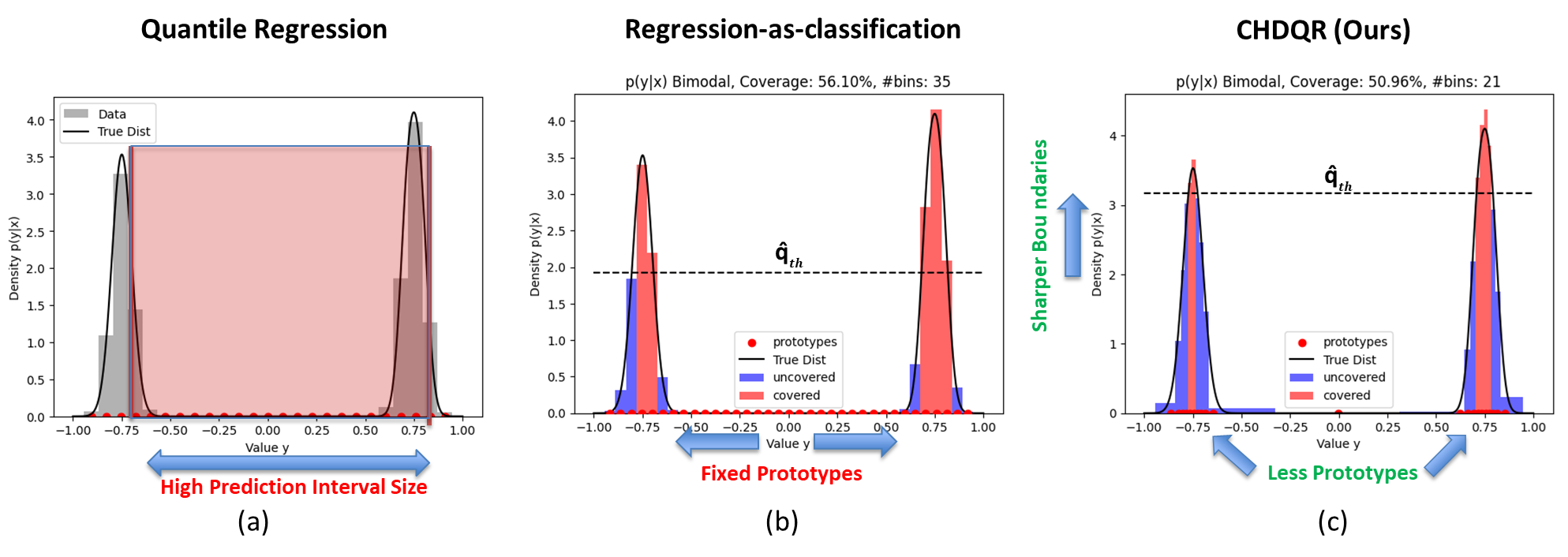}
    \caption{A simple demonstration for advantages of our method CHDQR. (a) Quantile Regression generates wide bands. (b) Regression-as-classification approach uses static bins, disregarding data distribution. (c) CHDQR dynamically updates prototypes toward data and discretizes dense regions with higher precision.}
    \label{fig:motivation_fig}
\end{figure}

To address these challenges, we propose a novel method called \textbf{Conformalized High-Density Quantile Regression (CHDQR)}, which integrates a scalable regression-as-classification (RasC) framework with dynamically adaptable prototypes for high-density region estimation. This approach introduces an adaptive prototype-based classification for high-density quantile regression, enabling prototype regions to update continuously based on the underlying data distribution. By combining RasC with dynamic prototypes, CHDQR enhances scalability and captures complex, non-convex high-density quantile regions, overcoming limitations of traditional convex quantile methods in handling multimodal and high-dimensional data distributions. As illustrated in Figure \ref{fig:motivation_fig}, CHDQR offers distinct advantages over previous methods: while quantile regression (QR) methods generate large prediction regions to maintain coverage, existing RasC methods are constrained by fixed quantization bin centers (prototypes) \cite{van2017neural}, limiting flexibility. CHDQR’s adaptive prototype approach addresses this gap, producing high-quality prediction regions. The main contributions of the paper can be summarized as follows: 
\begin{itemize}
    \item A scalable regression-as-classification method that allows quantile estimation to adaptively capture complex structures in high-dimensional target data.
    \item Dynamic prototypes that adjust based on the data, improving the flexibility and accuracy of quantile predictions, especially in non-convex and multimodal settings.
    \item A conformalized high-density region estimation framework that ensures reliable uncertainty estimates, providing robust prediction regions even in safety-critical applications.
\end{itemize}

The remainder of this paper is organized as follows: Section \ref{sec_method} presents all the details of the proposed CHDQR approach for multi-target regression problems. Section \ref{sec_exp} presents the exhaustive and comparative results. Finally, Section \ref{sec_conc} concludes the paper and outlines directions for future work.

\section{CHDQR Framework}
\label{sec_method}

This section describes the CHDQR \footnote{The code is available at \href{https://github.com/batuceng/max_quantile}{\color{magenta}https://github.com/batuceng/max\_quantile}}methodology used to develop a high-density quantile regression model for multi-targets. Motivated by the results of \cite{guha2024conformal}, we construct a RasC learning framework as presented in {Algorithm} \ref{Algo: Main_loop} but estimate probability densities rather than the probabilities. To overcome the exploiting prototype issue of this approach, we also develop a dynamic prototype method that is capable of adding and removing prototypes as outlined in {Algorithm} \ref{Algo: Add_Remove_Proto}.  The deployed conformalization procedure based on density estimation is summarized in {Algorithm} \ref{Algo: Conformal_Prediction}. In this section, we provide an in-depth explanation of these three primary components.

\input{algorithms/MainLoop.tex}

% \subsection{Preliminary} \label{Preliminary}
\subsection{Regression-as-Classification for density estimation} \label{CaR}

We assume having training dataset $\mathcal{D}_{tr} = \{(x, y)\}$ where $x$ denote the input variables and  $y \in \mathbb{R}^d$ the true multi-target values with $d$ as the number of targets. By quantizing $y$ into a finite set of point $\hat{y} \in C$, we can define the regression problem as a classification problem. We define the probability vector $q$ to label each prototype as a one-hot encoded vector and call this method hard quantization.
\begin{equation} \label{eq:hardq}
    q_i(y) = \begin{cases}
                1,  & \displaystyle\argmin_{i} \| y - c_i \| \\
                0,  & \text{otherwise}
                \end{cases}
    , \quad 
    \Sigma_i q_i = 1
\end{equation}
We can observe that we aim to quantize the continuous label $y$ by mapping it to the closest prototype from a set of learnable prototypes $C=\{c_1, c_2, \dots, c_K\}$, where $c_i \in \mathbb{R}^d$ and $K$ is the number of prototypes. The quantization process is defined as:
\begin{equation} \label{eq:quantization}
E_q =  \| y - \hat{y} \|
\quad \text{where} \quad
\hat{y} = \argmin_{c_i \in \{c_1, c_2, \dots, c_k\}} \| y - c_i \|
\end{equation}
where $E_q$ is the quantization error and $\hat{y}$ is the quantized approximation of $y$  \cite{guha2024conformal}.
% \textcolor{red}{[RasC yayınına atıf]}. 

One of the main drawbacks of regression-as-classification setting is the hard quantization given in Eq. \eqref{eq:hardq} as it enforces to assign each $y$ to its closest prototype $c_i$ by creating a set of probability vectors that are orthogonal to each other. In plain words, in the quantized space, all prototypes are equally distant from one another, regardless of their actual similarities or distances in the original space. Consequently, the Euclidean distance between any two quantization vectors becomes identical, failing to capture the nuanced relationships and inherent structures among the prototypes. To overcome this issue, we use soft quantization $q_i(y, \tau)$ as follows: 
\begin{equation} \label{eq:def_labeling}
    q_i(y,\tau) = \frac{exp({- \| y - c_i \|} / {\tau})}
                        {\sum_{j=1}^{K} exp({- \| y - c_j \|} / {\tau})}
    , \quad
    \lim_{\tau \to 0}q_i(y, \tau) = \begin{cases}
                1,  & \displaystyle\argmin_{i} \| y - c_i \| \\
                0,  & \text{otherwise}
                \end{cases}
\end{equation}
Note that, in theory, hard quantization is a special case of soft quantization with $\tau=0$.

As we aim to develop a regression model based on densities, we define the region $R_i$ as the area around prototype $c_i$ where $c_i$ is the closest prototype to the point. Using this definition, we quantize the output space $y$ into subspaces where each region is represented as a class in $q_i$ label vector.
\begin{equation} \label{eq:def_region}
R_i = \{ s \in \mathbb{R}^d \mid \| s - c_i \| \leq \| s - c_j \| \ \forall j \neq i \} 
\end{equation}
Next, we define the probability of each region based on its area and density. In the equidistant binning method, probability is linearly proportional to density. However, our quantized regions are not equidistant, so we must calculate both the area size and the density explicitly.

\begin{equation} \label{eq:pdf_cont}
    P(y \in R_i) = \oint_{R_i} p_i \,dV
\end{equation}
Here, we express that the probability of $y$ being inside the region $R_i$ of prototype $c_i$ is equivalent to the integration over the volumetric space with a probability density function $p_i$. We discretize this continuous integration using the midpoint rule and assume that each prototype has a uniform-valued probability density function (pdf) inside its region. Therefore, the probability of a region can be calculated using the pdf value $p_i$ and the area size $A_i$.

\begin{equation} \label{eq:pdf_disc}
    P(y \in R_i) = p_i \cdot A_i,
    \quad
    A_i = \oint_{R_i} \,dV ,
    \quad
    p_i \in \mathbb{R}_{\geq 0}
\end{equation}

The main challenge in computing Eq. \eqref{eq:pdf_disc} is the fact that the true probabilities and the region areas are not directly observable. Therefore, we propose two key approximations:  (i) we compute the boundaries and the size of each region using the Voronoi algorithm \cite{Voronoi1908Nouvelles,aurenhammer1991voronoi}, assuming that these regions accurately reflect the decision boundaries of the prototypes, and (ii) we model the log-probability density function using a neural network $f(x; \theta)$, where $\theta$ are the parameters of the network. The neural network approximates the conditional log-density $p_i(x)$ for each region.
\begin{equation} \label{eq:approximators}
Voronoi(C)_i = A_i,
\quad
f(x;\theta)_i = \log p_i(x)
\end{equation}
% Here, we condition our model on the input variable $x$ since we have the training data samples $D_{tr} = \{(x, y)\}$. Overall, 
We condition $f(x; \theta)$ through the following conditional probability distribution:
\begin{equation} \label{eq:softmax_pred}
    P[y \in R_i \mid x] = p_i(x) \cdot A_i \implies
    \hat{P}[y \in R_i \mid x] = \frac{\exp(f(x;\theta)_i + \log A_i)}
                                {\sum_{j} \exp(f(x;\theta)_j + \log A_j)}
\end{equation}
The softmax-like formulation in Eq. \eqref{eq:softmax_pred} ensures that the conditional probabilities for all regions sum to one, accounting for both the region area and the learned log-probability density function. Additionally, we note that learning $\log p_i(x)$ instead of the direct value ensures non-negativity.

%% Losses
\subsection{Developing the Loss function} \label{Losses}
As we suggested, we want to learn both the quantization regions $R_i$ parameterized by prototype set $\mathcal{C}$ and the log-likelihood estimator $f(x;\theta)$ parameterized by $\theta$. In this context, we propose the following composite loss function $\mathcal{L}$ which is a weighted combination of the cross-entropy loss $\mathcal{L}_{CE}$, minimum distance loss $\mathcal{L}_{q}$, and repulsion loss $\mathcal{L}_{rep}$:
\begin{equation} \label{eq:total_loss}
    (\theta^*, C^*) = \argmin_{\theta, C} \mathcal{L}
                        \text{, } \qquad
                        \mathcal{L} = \mathcal{L}_{CE} + 
                        \lambda_{q} \cdot \mathcal{L}_{q} + 
                        \lambda_{rep} \cdot \mathcal{L}_{rep}
\end{equation}
Here,
\begin{itemize}
    \item $\mathcal{L}_{CE}$ is the cross-entropy loss that is included to optimize our log-likelihood estimator. 
    \begin{equation} \label{eq:crossentropyloss}
    \mathcal{L}_{CE} = \sum_i q_i(y,\tau) \cdot \log \hat{P}[y \in R_i \mid x], 
    \quad 
    \log \hat{P}[y \in R_i \mid x] = f(x;\theta) + \log A_i
\end{equation}
    We use labeling probability vector $q$ defined in Eq. \eqref{eq:def_labeling} and predicted probability vector $\hat{P}$ defined in equation Eq. \eqref{eq:softmax_pred}.
    \item $\mathcal{L}_{q}$ is the distance between regression target $y$ and assigned prototype $c_i$ is included as
    \begin{equation} \label{eq:mindistloss}
    \mathcal{L}_{q} = \min_{c_i \in \{C\}} \| y - c_i \|
    \end{equation}
    since we want to optimize the prototypes to minimize the quantization error $E_q$ defined in equation Eq. \eqref{eq:quantization}. Through the inclusion of $\mathcal{L}_{q}$, we update assigned prototypes towards the $y$ points and simulate a k-means clustering-like algorithm.
    
    \item 
    $\mathcal{L}_{rep}$ is the repulsion loss that aims to prevent prototypes from collapsing to single points for high-density regions.
    \begin{equation} \label{eq:repulsionloss}
    \mathcal{L}_{rep} = \sum_{i \neq j} \max(0, \|c_i - c_j\| - \delta_{rep})
\end{equation}
    The aim of the inclusion of $\mathcal{L}_{rep}$ is to keep the granularity of region size $A_i$ over a desired minimum size. Note that, while $\mathcal{L}_{q}$ closes the gap between the prototype and the target value by attraction, $\mathcal{L}_{rep}$ distracts prototypes from each other if they are closer than a desirable threshold $\delta_{rep}$.
\item $\lambda_{q}$ and $\lambda_{rep}$ are hyperparameters that control the trade-off between minimizing the quantization error and maintaining separation between prototypes.
\end{itemize}

\begin{figure}[t]
    \centering   \includegraphics[width=0.99\textwidth]{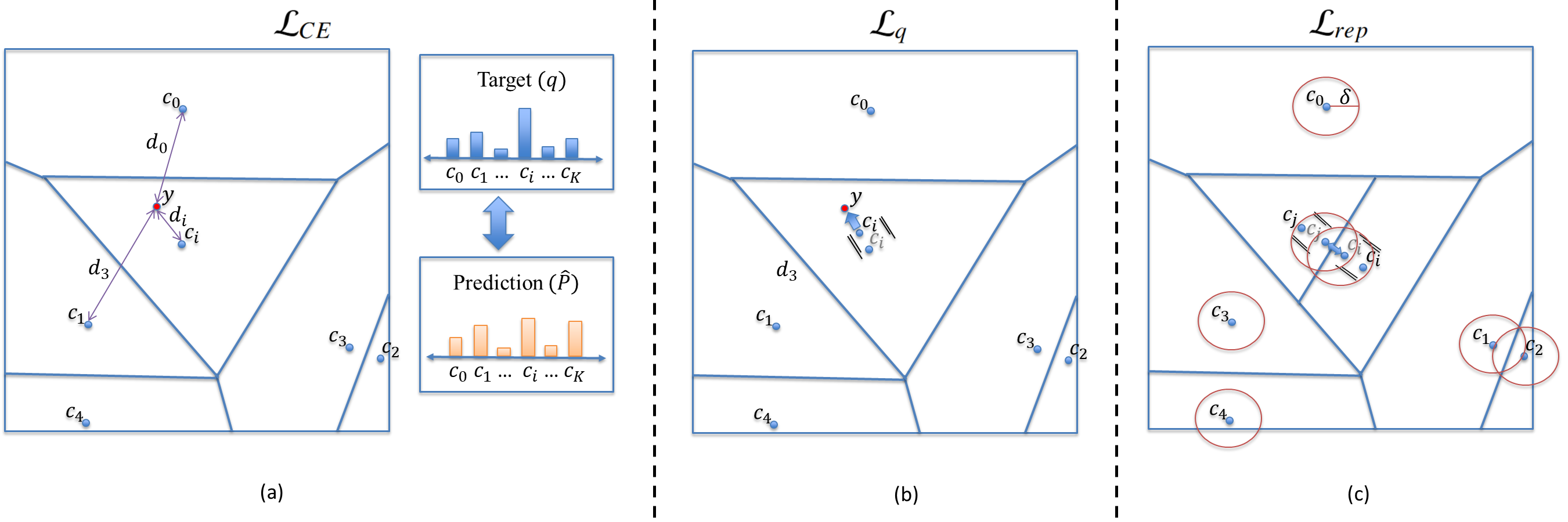}
    \caption{Êffect of different Loss functions. (a) Cross-entropy loss closes the distance between target distribution $q$ and the prediction $\hat{P}$. (b) Shows how $\mathbb{L}_{q}$ moves closest prototype toward $y$. (c) Shows $\delta$ threshold of $\mathbb{L}_{rep}$ and how it pushes prototypes from each other within range.}
    \label{fig:losses}
\end{figure}

In Fig. \ref{fig:losses}, we demonstrated how each component of $\mathcal{L}$ affects the training. The overall training procedure alternates between updating the log-likelihood estimator $\theta$ and the prototype set $\mathcal{C}$ by minimizing $\mathcal{L}$, ensuring both accurate probability predictions and well-separated quantization regions.

% In summary, we optimize the parameters $\theta$ (of the log-likelihood estimator) and the prototype set $\mathcal{C}$ by minimizing the total loss, which is a weighted combination of the cross-entropy loss $\mathcal{L}_{CE}$, minimum distance loss $\mathcal{L}_{q}$, and repulsion loss $\mathcal{L}_{rep}$. 
% \begin{equation} \label{eq:total_loss}
%     (\theta^*, C^*) = \argmin_{\theta, C} \mathcal{L}
%                         \text{, } \qquad
%                         \mathcal{L} = \mathcal{L}_{CE} + 
%                         \lambda_{q} \cdot \mathcal{L}_{q} + 
%                         \lambda_{rep} \cdot \mathcal{L}_{rep}
% \end{equation}

\input{algorithms/AddRemoveProto.tex}

\subsection{Dynamic Number of Prototypes} \label{dynamic}

We define a dynamic mechanism for managing the number of prototypes $K$, allowing the model to adaptively adjust to an optimal value, $K^*$. Rather than keeping $K$ fixed, we introduce an algorithm that adds new prototypes in high-density regions and removes underutilized ones from sparse regions. We call these algorithms \textit{AddProto} and \textit{RemoveProto} within the pseudo-code shown in Algorithm \ref{Algo: Add_Remove_Proto}.

We first define the normalized usage of each prototype $c_i$ to determine how often it is utilized during training. The usage of a prototype is measured based on how frequently it is assigned to target points in the dataset $D_{tr}$. For each prototype $c_i$, the usage function $U(c_i)$ is defined as:

\begin{equation} \label{eq:usage}
    U(c_i) = \frac{1}{|\mathcal{D}_{tr}|} \sum_{y \in \mathcal{D}_{tr}} q_i(y,\tau)
\end{equation}
where $q_i(y,\tau)$ represents the soft quantization of the target point $y$ with respect to prototype $c_i$, as defined in Eq. \eqref{eq:def_labeling}. The summation is performed over the entire training set $D_{tr}$, and the result is normalized by the size of the training set, $|D_{tr}|$. This gives us a measure of how frequently each prototype $c_i$ is utilized relative to the other prototypes.
If the usage of a prototype falls below the threshold $\delta_{\text{del}}$, the prototype is considered underutilized and is removed:

\begin{equation} \label{eq:delete_proto}
    C \longleftarrow C - \{ c_i \} \quad \text{if }  U(c_i) \leq \delta_{\text{del}}
\end{equation}
This ensures that prototypes that do not contribute significantly to representing the regression space are discarded, helping to prevent overfitting in sparse regions. After removing a prototype, the log-likelihood estimator $f(x;\theta)$ must be updated:

\begin{equation}
    f^{(-1)}(x;\theta) = Wx + b
\end{equation}
Since the output size of this network is equal to the number of prototypes $K$, we remove the corresponding row of weights associated with the discarded prototype $c_i$ from the final linear layer. The overall probability distribution is then updated using the softmax prediction from Eq. \eqref{eq:softmax_pred}.

\begin{equation} \label{eq:delete_weight}
    W \longleftarrow W - \{ w_i \}, \quad b \longleftarrow b - \{ b_i \}
\end{equation}

Similarly, if the usage of a prototype exceeds the threshold $\delta_{\text{add}}$, a new prototype $c_{K+1}$ is added at a small $\epsilon \sim \mathcal{N}(0, \sigma^2)$ distance from the existing prototype $c_i$.

\begin{equation} \label{eq:add_proto}
    C \longleftarrow \text{concat}(C, \{ c_{K+1} \}) \quad \text{if } U(c_i) \geq \delta_{\text{add}}
    , \quad
    c_{K+1} \sim \mathcal{N}(c_i, \sigma^2)
\end{equation}
This rule ensures that high-density regions where a prototype is frequently assigned will have greater representation, improving the overall accuracy of the model.

\begin{equation} \label{eq:add_weight}
    W \longleftarrow \text{concat}(W, \{ w_i \}), \quad b \longleftarrow \text{concat}(b , \{ b_i \})
\end{equation}
As we have done in the deleting procedure, we also update the weights and biases accordingly. We assume that the newly added prototype, which is at a small $\epsilon$ distance from the original one, has approximately the same log-likelihood since the distribution is assumed to be uniform in region $R_i$.

% The dynamic prototype management algorithm, combining the above rules, allows the model to adaptively adjust the number of prototypes during training. This leads to a more efficient and flexible representation of the regression space, resulting in an optimal number of prototypes $K^*$.

By iterating over these steps during the training process, the model can maintain a balanced and adaptive prototype set, ensuring that the prototypes are well-distributed and relevant to the data distribution.

\subsection{Conformalized High Density Quantile Regression} \label{CHDQR}

\input{algorithms/Conformalize.tex}
% 'High Density Quantile': Uzayın (1-alpha) en az olasılıklı alanını kapmalayan ve en yüksek density'li yerlerin oluşturduğu setlerin union'ının en küçüğü. Örneğin Gamma(0.9) uzayın olasılık olarak %90'ını temsil eder ancak dışarıdaki hiçbir noktanın density'si içeridekilerden büyük değildir... 
We define high-density quantile region $\Gamma_{1 - \alpha}(x)$ as the smallest set of regions \( R_i \), sorted by their density estimates \( p(x) \), such that the cumulative probability reaches at least \( 1 - \alpha \). In other words, \( \Gamma_{1 - \alpha}(x) \) is the union of regions with the highest density.  Mathematically, we describe $\Gamma_{1 - \alpha}(x)$ as follows:
\begin{equation} \label{eq:high_density_quantile}
\Gamma_{1 - \alpha}(x) = \sup \left\{ R_{\pi_i} : \sum_{i=1}^{r} [y \in R_{\pi_i}] \geq 1 - \alpha, \quad p(x)_{\pi_{i}} \leq p(x)_{\pi_{i+1}} \ \forall i \right\}
\end{equation}
% 'pi': Sorted density'lerin indisleri
where, \( \pi(x) \) represents the indices of regions sorted by their density estimates \( p(x) \), ensuring that any region outside \( \Gamma_{1 - \alpha}(x) \) has an equal or lesser density than those inside. However, since directly estimating the highest density region to cover a space with probability \( 1 - \alpha \) is not feasible, we introduce a cumulative density function \( \rho_K(x) \) to define the total probability of a high-density quantile \( \Gamma_{1-\hat{q}}(x) \), as 
\begin{equation} \label{eq:cumulative_probs}
    \rho_K(x) := 
    {P}(y \in \Gamma_{1 - \hat{q}}(x) \mid x) 
    = \sum_{i=1}^{K} \hat{P}[y \in R_{\pi_i} \mid x]
\end{equation}
which consist of the union of the top \( K \) dense regions. The probability of this set is an arbitrary value $1-\hat{q}$.

To calibrate our algorithm, we define the following nonconformity score function $S(y_{cal}, x_{cal})$, which is similar to the previously defined  $\rho_K(x)$: 
\begin{equation} \label{eq:nonconf_score}
S(y_{cal}, x_{cal}) := \rho_r(x_{cal}) = \sum_{i=1}^{r} \hat{P}[y_{cal} \in R_{\pi_i} \mid x_{cal}] \quad \text{, where } \quad r:y_{cal} \in R_r
\end{equation}
Here we set $K=r$, where $r$ is the index of the region that includes true value $y_{cal}$. 
This approach allows for calculating the probability values for the smallest high-density regions that include our $y_{cal}$ values.

For a calibration dataset \( \mathcal{D}_{cal} \), the nonconformity score \( S(y_{cal}, x_{cal}) \) is computed by summing the probabilities \( \hat{P} \), where the sorting order is determined by the log-density estimates \( f(x; \theta) \). The quantile threshold \( \hat{q}_{\alpha} \) is defined as:

\begin{equation} \label{eq:quantile_qhat}
\hat{q}_{\alpha} = \text{quantile}(S(y_{cal}, x_{cal}) \mid \mathcal{D}_{cal}, \lceil 1 - \alpha + \frac{1}{|\mathcal{D}_{cal}| + 1} \rceil)
\end{equation}
% Bu denklem APS Makalesinde (Conformal Classification) açıklanıp kanıtlanıyor, ama biz nasıl yapabiliriz emin değilim...
which ensures the conformal coverage guarantee:
\begin{equation} \label{eq:coverage_guarantee}
1 - \alpha \leq P\left( Y_{n+1} \in \Gamma_{\hat{q}_{\alpha}}(X_{n+1}, \hat{q}_{\alpha}) \right) \leq 1 - \alpha + \frac{1}{|\mathcal{D}_{cal}| + 1}
\end{equation}

In the prediction step, for a test sample  $(x_{test}, y_{test})\in \mathcal{D}_{test}$, the prediction set \( \Gamma_{\hat{q}_{\alpha}}(x_{test}) \) includes all regions where the cumulative sum of probabilities remains below the threshold \( \hat{q}_{\alpha} \):
% Gamma seti için subscript'e ne yazılmalı? (1-alfa) mı yoksa qhat_alpha mı? Conformalized olduğundan dolayı yıldız falan mı koymalıyız?
\begin{equation} \label{eq:conformal_prediction}
\Gamma_{\hat{q}_{\alpha}}(x_{test}) = \left\{ R_{\pi_i(x_{test})} : \sum_{i=1}^{r} \hat{P}[y \in R_{\pi_i(x_{test})} \mid x_{test}] \leq \hat{q}_{\alpha} \right\}
\end{equation}
This process ensures that the prediction set is composed of regions with the highest density while maintaining the required coverage level.

\section{Experiments} \label{sec_exp}

In this section, we analyze the performance of the proposed methodology on synthetic and benchmark datasets. We also analyze the robustness of the proposed against outliers. 

In the experiments, we train and evaluate two versions of our proposed method: CHDQR and CHDQR-Dynamic. CHDQR utilizes learnable prototypes with a fixed number of cluster centers, while CHDQR-Dynamic dynamically adds and removes cluster centers based on the data distribution as shown in Algorithm \ref{Algo: Add_Remove_Proto}. We compare CHDQR and CHDQR-Dynamic by: 
\begin{itemize}
    \item reimplementing the R2CCP approach \cite{guha2024conformal}, labeled as GRID, and ensure compatibility with multi-target data, as the original implementation was limited to 1D data. We distributed the prototypes across a grid space with 50 bins per dimension, consistent with the original method. Unlike R2CCP, we employed standard $\mathcal{L}_{CE}$ loss with a soft labeling technique  $q(y,\tau)$  instead of the original loss proposed in the paper, as the latter did not yield the desired results on 2D data in our initial experiments. In our setup, the GRID consists of static, non-learnable cluster centers. A final difference between the GRID method and the original R2CCP is that GRID defines discrete probabilities and does not employ interpolation to estimate a continuous distribution, as this is not applicable for higher dimensions.
    \item implementing the CQR \cite{romano2019conformalized} algorithm trained via the pinball loss. In our adaptation for the multi-target task, we regressed each quantile for each target dimension individually and used the cross-product of these regions as the prediction. However, a drawback of this approach is that pinball loss operates on the marginal distributions of the target values, which caused the method to perform poorly in the multi-target experiments.
\end{itemize}

In all our experiments, we focused exclusively on quantile prediction, excluding point-wise predictions. For dataset splitting, we allocated 80\% for training, 10\% for calibration, and 10\% for testing. To ensure robustness, we repeated each training process with 10 different random seeds for each dataset. All methods were tested on these pre-saved splits to maintain consistency across experiments.

\subsection{Metrics}

%As a metric we are used to method used in \cite{quan2014short} which they are names as PI Coverage Probability (PICP), and PI Normalized Averaged Width (PINAW) and extended them to multidimensional target. 

To assess the quantitative quality of the prediction regions, we use metrics inspired by PI Coverage Probability (PICP) and PI Normalized Average Width (PINAW) \cite{quan2014short}. These original metrics, while effective, are limited to 1D data and convex sets. Building on their definitions, we introduce Coverage and PINAW for multidimensional data.

The Coverage metric, defined in Eq. \eqref{eq:covarage_metric}, represents the proportion of test samples for which the true label falls within the predicted region:
\begin{equation} \label{eq:covarage_metric}
\text{Coverage} = \frac{1}{|\mathcal{D}{\text{test}}|} \sum{(x, y) \in \mathcal{D}_{\text{test}}} \mathds{1}[y \in \Gamma(x)]
\end{equation}
The size of the prediction region, referred to as PINAW (for consistency with the original metric in \cite{quan2014short}), is calculated by summing the areas of all prototypes included in the prediction set, as shown in Eq. \eqref{eq:pinaw_metric}:
\begin{equation} \label{eq:pinaw_metric}
\text{PINAW} = \frac{1}{|\mathcal{D}{\text{test}}|} \sum{(x, y) \in \mathcal{D}{\text{test}}} \sum{R_i \in \Gamma(x)} A_i
\end{equation}
where $A_i$  denotes the area of the Voronoi region associated with prototype  $c_i$. Both metrics are averaged over the entire test set  $\mathcal{D}_{\text{test}}$  to ensure consistency and reliability in performance evaluation.
\subsection{1D Dataset Results} 

In this section, we present results from three sets of experiments conducted on seven datasets: {Bike}\cite{bike_sharing_275}, {Concrete}\cite{concrete_compressive_strength_165}, {MEPS19}, {MEPS20}, {MEPS21} \cite{meps}, {Wine} \cite{wine_quality_186}, and a synthetic toy dataset named {Uncond1d}. The experiments are summarized in three tables, each corresponding to different target coverage levels: 90\%, 50\%, and 10\%. For the real datasets, we evaluated the methods based on the coverage and size of the prediction region (i.e. PINAW) to demonstrate the effectiveness of the proposed approaches.
% Dataset'in unconditional olduğu belirtilmeli, input x=0 always
To further validate the performance of the methods, {Uncond1d} is designed to be bimodal, generated by combining two distinct normal distributions. Specifically, the probability density function (PDF) for {Uncond1d} is given at \ref{app:Sythetic}. This design introduces controlled complexity, enabling us to analyze model performance on non-trivial data distributions.

\input{tables/table_1d_90}

Table \ref{tab:1d-comparison-0.9} shows that all methods achieved the desired coverage values with minor deviations, which is expected, particularly for 90\% coverage—a relatively attainable target for such methods. However, examining the PINAW values reveals clear differences in performance. For the Bike, Concrete, and Wine datasets, CHDQR-Dynamic predicted the smallest prediction region while still meeting the coverage requirement, indicating its effectiveness in refining prediction regions.

\begin{figure}[ht]
    \centering
    \includegraphics[width=0.99\textwidth]{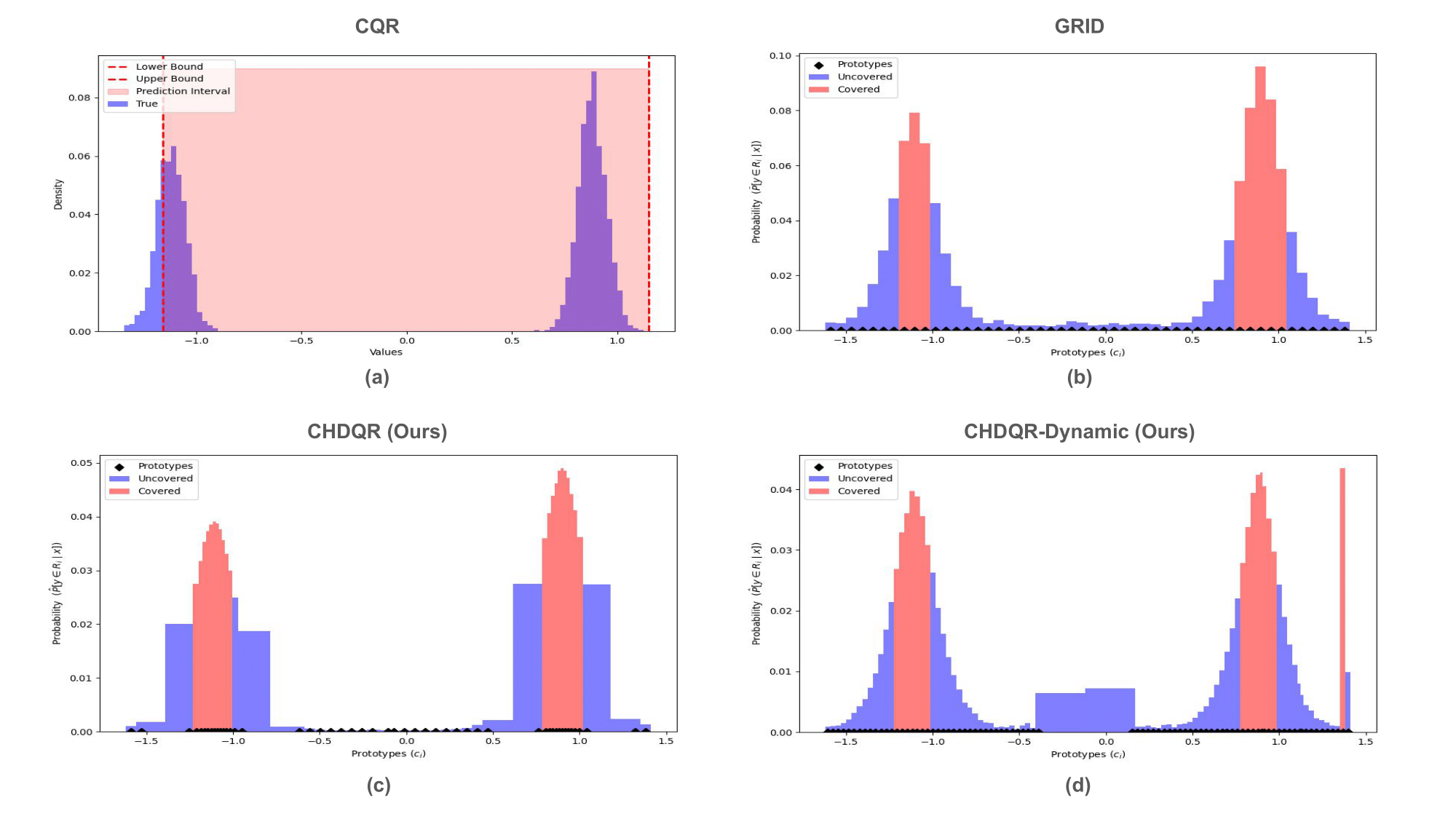}
    \caption{Visualization of {Uncond1d} Dataset for 90\% Coverage}
    \label{fig:1d_90}
\end{figure}

On the other hand, increasing the number of cluster centers in CHDQR-Dynamic led to some overfitting, as seen in Figure \ref{fig:1d_90}, with high probability predictions appearing in regions with sparse or no data. This caused CHDQR to perform better than CHDQR-Dynamic in the {Uncond1d} dataset, though CHDQR-Dynamic still outperformed CQR and GRID overall.

In contrast, CQR showed the best PINAW values for the MEPS datasets, likely due to the narrow, concentrated distribution of MEPS data, which favored CQR’s simpler, static approach. This highlights CHDQR-Dynamic’s advantage in handling complex distributions, while CQR remains effective for simpler datasets.

Overall, our method demonstrated superior performance, particularly for more complex or bimodal structures. CHDQR-Dynamic’s adaptability in efficiently splitting the space results in smaller and more accurate prediction regions, as illustrated in Figure \ref{fig:1d_90}, demonstrating its strength in tackling diverse data distributions.

\input{tables/table_1d_50}

Table \ref{tab:1d-comparison-0.5} shows that all methods successfully reached the 50\% target coverage across different datasets, with only slight deviations. Achieving precise coverage at this lower level is more challenging due to the narrower prediction region required, making these results particularly noteworthy.

Examining the PINAW values reveals significant differences among the methods. CHDQR-Dynamic consistently produced the smallest prediction region across most datasets, including Bike, Concrete, Wine, and Uncond1d, while still meeting the coverage requirements. This indicates that CHDQR-Dynamic is highly effective at refining prediction regions, providing more precise and informative predictions.

\begin{figure}[ht]
\centering
\includegraphics[width=0.99\textwidth]{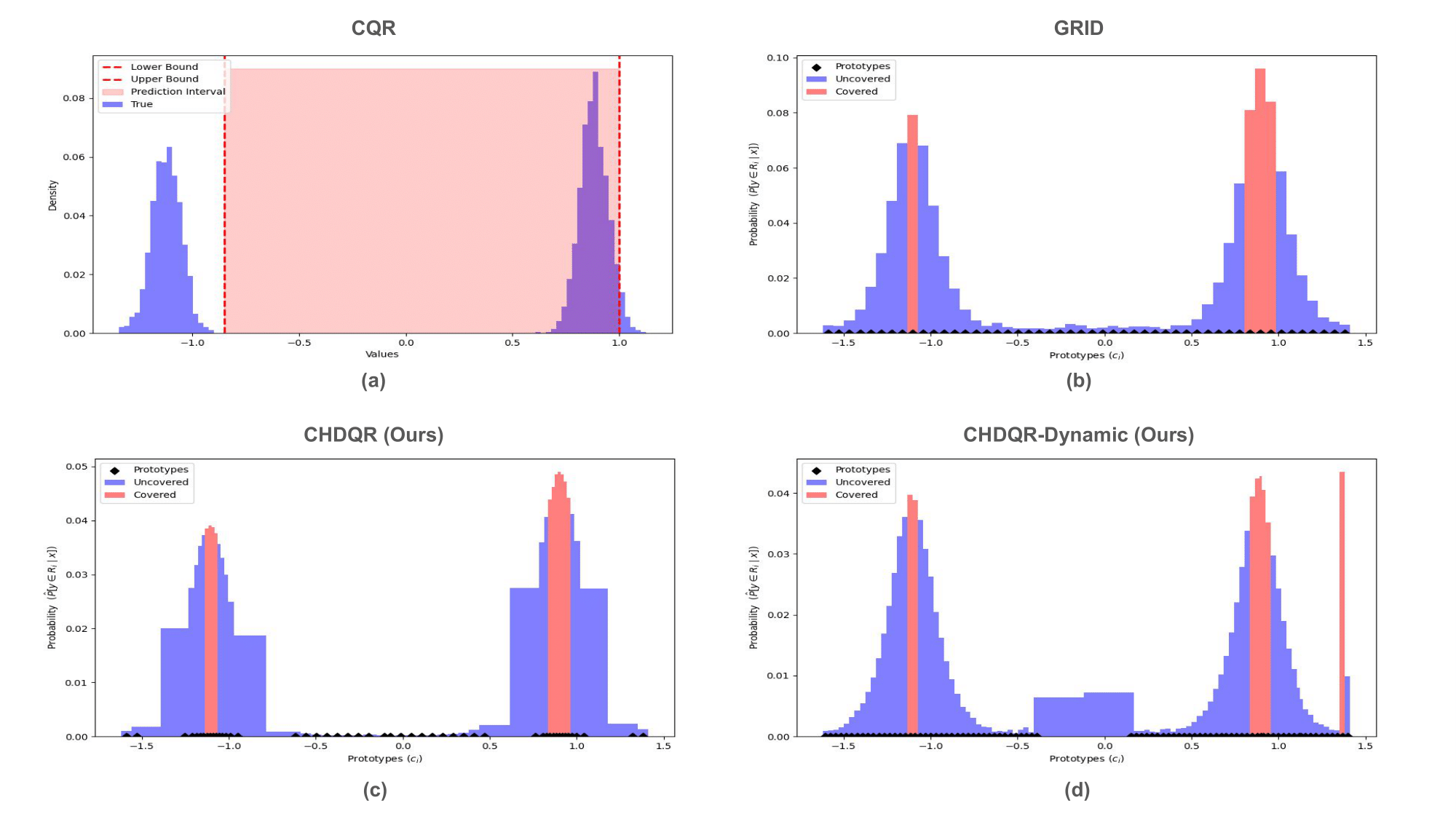}
\caption{Visualization of Unconditional1d Dataset for 50\% Coverage}
\label{fig:1d_50}
\end{figure}

\input{tables/table_1d_10}

In the MEPS datasets, CHDQR-Dynamic outperformed the other methods in MEPS19 and MEPS21 by achieving lower PINAW values. Although CQR slightly outperformed CHDQR-Dynamic in MEPS20, the difference was minimal. %This suggests that while CHDQR-Dynamic excels in handling complex and diverse data distributions, CQR’s simpler approach remains competitive in datasets with more concentrated distributions.

In the CHDQR method, the density of learnable cluster centers increases in regions where data samples are more concentrated. As shown in Figure \ref{fig:1d_50}, this allows us to predict more precise coverage levels with narrower regions. Consequently, our model performs better at lower coverage levels.
\par

\begin{figure}[h!]
\centering
\includegraphics[width=0.99\textwidth]{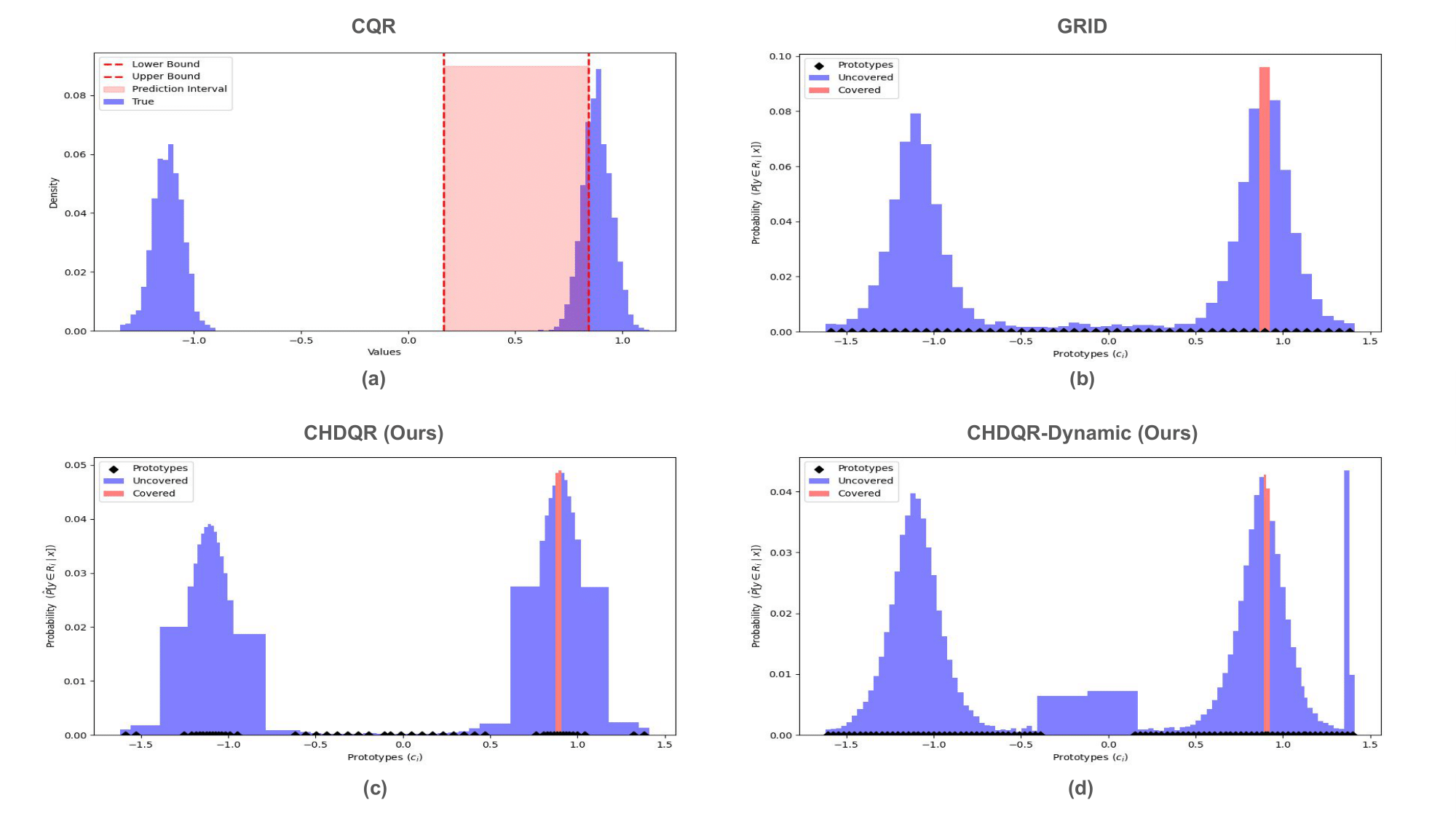}
\caption{Visualization of Unconditional1d Dataset for 10\% Coverage}
\label{fig:1d_10}
\end{figure}

Table \ref{tab:1d-comparison-0.1} shows that CHDQR-Dynamic consistently produced the smallest prediction regions across all datasets, demonstrating its ability to create more precise and accurate clusters, enabling narrower regions for small target coverages. As shown in Figure \ref{fig:1d_10}, the covered regions clearly indicate that CHDQR-Dynamic achieves the most compact area while successfully covering 10\% of the data.

\subsection{2d Dataset Results}

In this section, we evaluate the performance of different methods on 2D datasets, where the higher output dimension presents a more complex challenge. We anticipate that our CHDQR-Dynamic method will excel in these scenarios due to its ability to efficiently refine and partition the data space. In contrast, CQR, which is trained using marginal  Y  values, struggles to achieve the desired coverage in multi-dimensional cases.

To demonstrate these capabilities, we used the {Energy} dataset \cite{energy_efficiency_242} and our own {Uncond2d} dataset. The {Uncond2d} dataset consists of three Gaussian distributions with varying means and covariance matrices, combined to create a challenging, multi-modal data structure. Data distributions and more information are given in \ref{app:Sythetic}.

\input{tables/table_2d}

\begin{figure}[h!]\label{fig:2coverage_levels}
    \centering
    \begin{subfigure}[b]{0.49\textwidth}
        \centering
        \includegraphics[width=\textwidth]{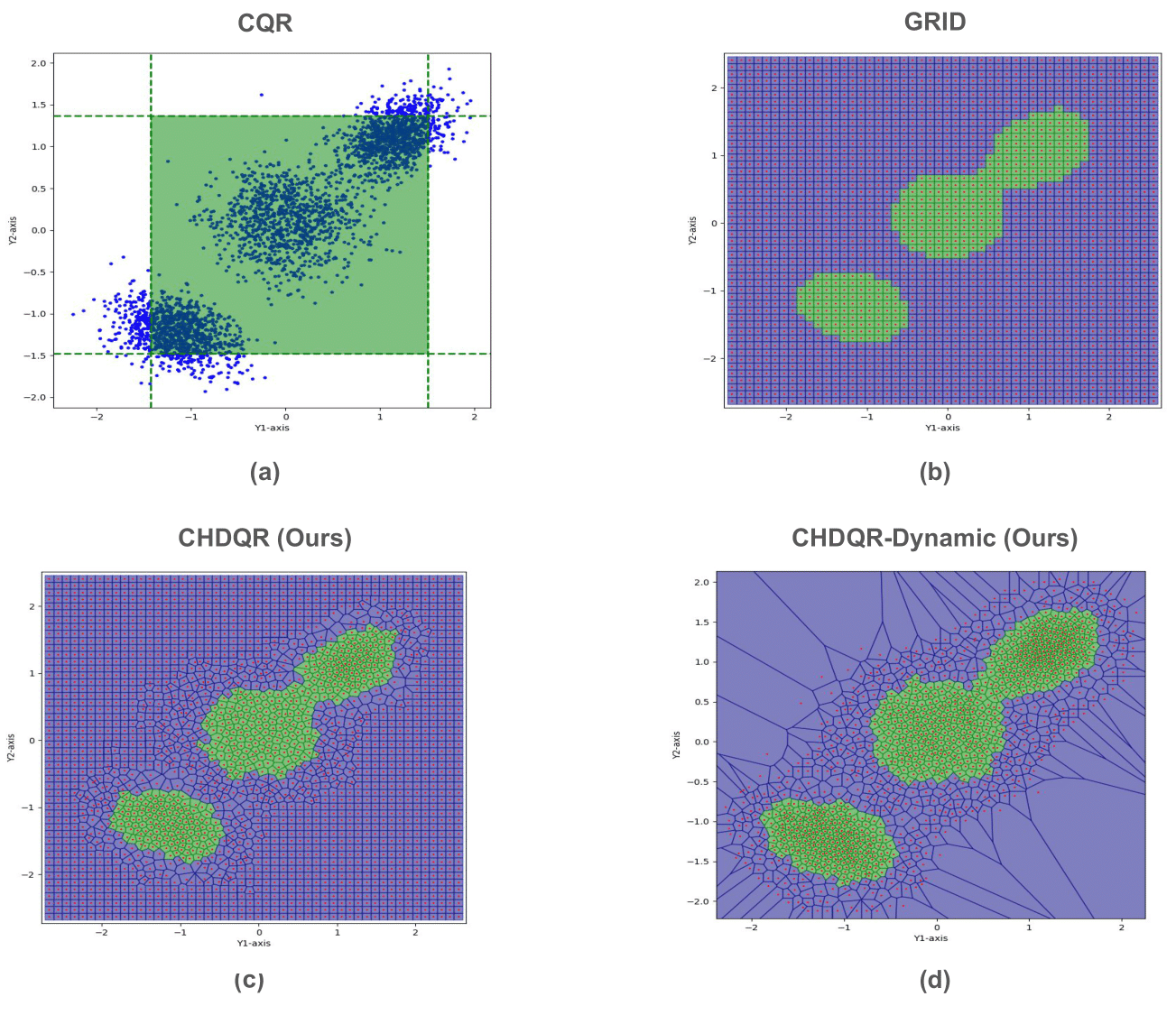}
        \caption{90\% Coverage}
        \label{fig:2d_0.9}
    \end{subfigure}
    \hfill
    \begin{subfigure}[b]{0.49\textwidth}
        \centering
        \includegraphics[width=\textwidth]{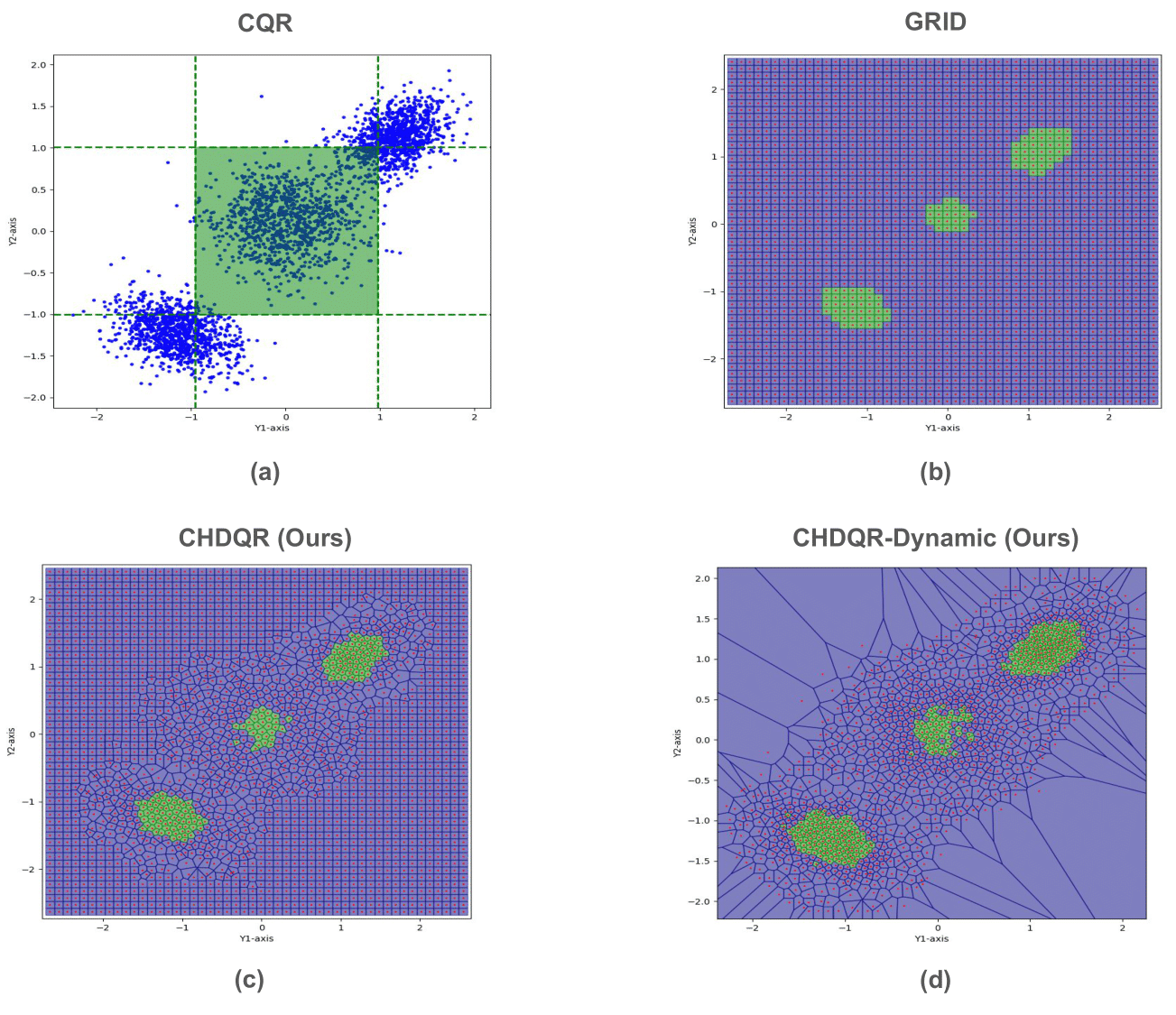}
        \caption{50\% Coverage}
        \label{fig:2d_0.5}
    \end{subfigure}
    \hfill
    \begin{subfigure}[b]{0.55\textwidth}
        \centering
        \includegraphics[width=\textwidth]{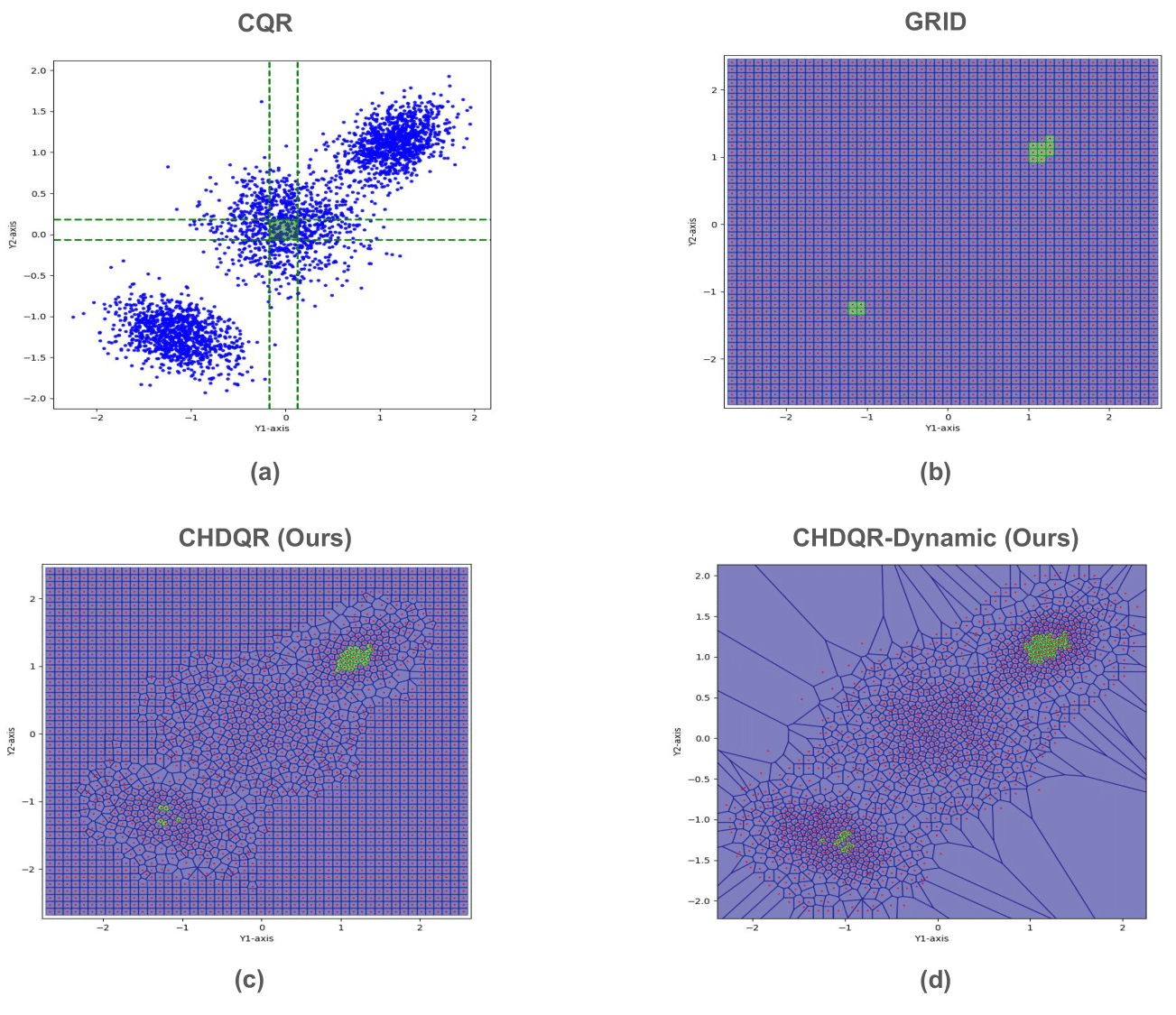}
        \caption{10\% Coverage}
        \label{fig:2d_0.1}
    \end{subfigure}
    \caption{Visualization of Uncond2d Dataset at Different Coverage Levels}
    \label{fig:2d_coverage_levels}
\end{figure}

In Table \ref{tab:combined-comparison}, for 90\% target coverage, the GRID, CHDQR and CHDQR-Dynamic methods achieved the desired coverage consistently across both datasets, while CQR struggled to reach this target, especially in the {Uncond2d} dataset. The PINAW values indicate that CHDQR-Dynamic produced the narrowest prediction regions, suggesting its effectiveness in refining and adjusting prediction regions with fewer clusters.

% Yüzde 50 için table 4te yorum yazamadım 

With nearly half the cluster centers compared to GRID and CHDQR, the CHDQR-Dynamic method achieves superior results as can also be seen in Figures \ref{fig:2d_0.9}, \ref{fig:2d_0.5} and \ref{fig:2d_0.1}, demonstrating both efficiency and precision. As dimensions increase, the required number of clusters typically scales exponentially as $n^d$, making the efficient use of clusters crucial in high-dimensional spaces. By focusing clusters only where needed, CHDQR-Dynamic significantly reduces the computational complexity for continuous region prediction, enabling high performance without the cost of dense clustering. This sparsity and adaptability make CHDQR-Dynamic especially valuable for complex, multi-dimensional data, where managing computational load is essential.

Overall, the results confirm CHDQR-Dynamic’s superior performance in maintaining accurate coverage with efficient clustering and narrower prediction regions, making it well-suited for handling diverse and complex data distributions.

\subsection{Robustness to outliers}

Our model demonstrates robust performance, particularly in handling diverse data with embedded outliers, due to its ability to dynamically adapt to local regions within the data space. To assess this robustness, we introduced outliers to our synthetic {Uncond2d} dataset, in two separate experiments.

In the first experiment, we added 100 samples from two distinct Gaussian distributions positioned far from the original data distribution, representing approximately $\sim0.66\%$ of the total dataset. We then plotted the cluster distributions and calculated our metrics. In the second experiment, we increased the number of outlier samples to 1000 from each of the two Gaussian distributions, making up $\sim6.6\%$ of the total dataset. At this higher density, these additional samples began to merge with the main distribution, forming a distinct substructure rather than remaining isolated outliers

These experiments highlight our model’s adaptability and robustness across different data configurations, effectively handling both isolated and clustered outliers.

\input{tables/table_2d_outlier_90}

\begin{figure}[ht]
    \centering
    \begin{subfigure}[b]{0.49\textwidth}
        \centering
        \includegraphics[width=\textwidth]{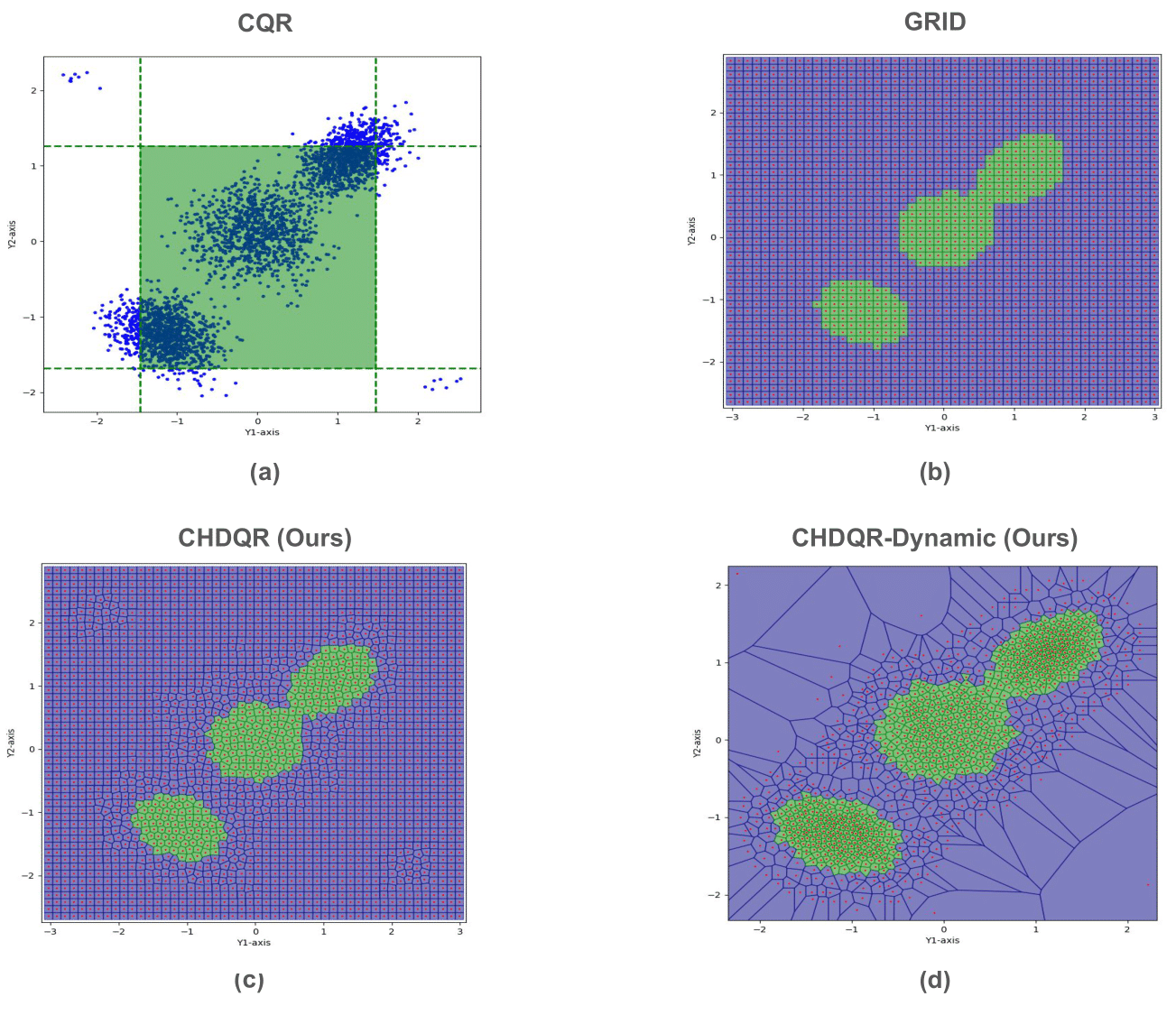}
        \caption{Uncond2d Outlier (0.6\%) with 90\% Coverage}
        \label{fig:2d_outlier_0.6}
    \end{subfigure}
    \hfill
    \begin{subfigure}[b]{0.49\textwidth}
        \centering
        \includegraphics[width=\textwidth]{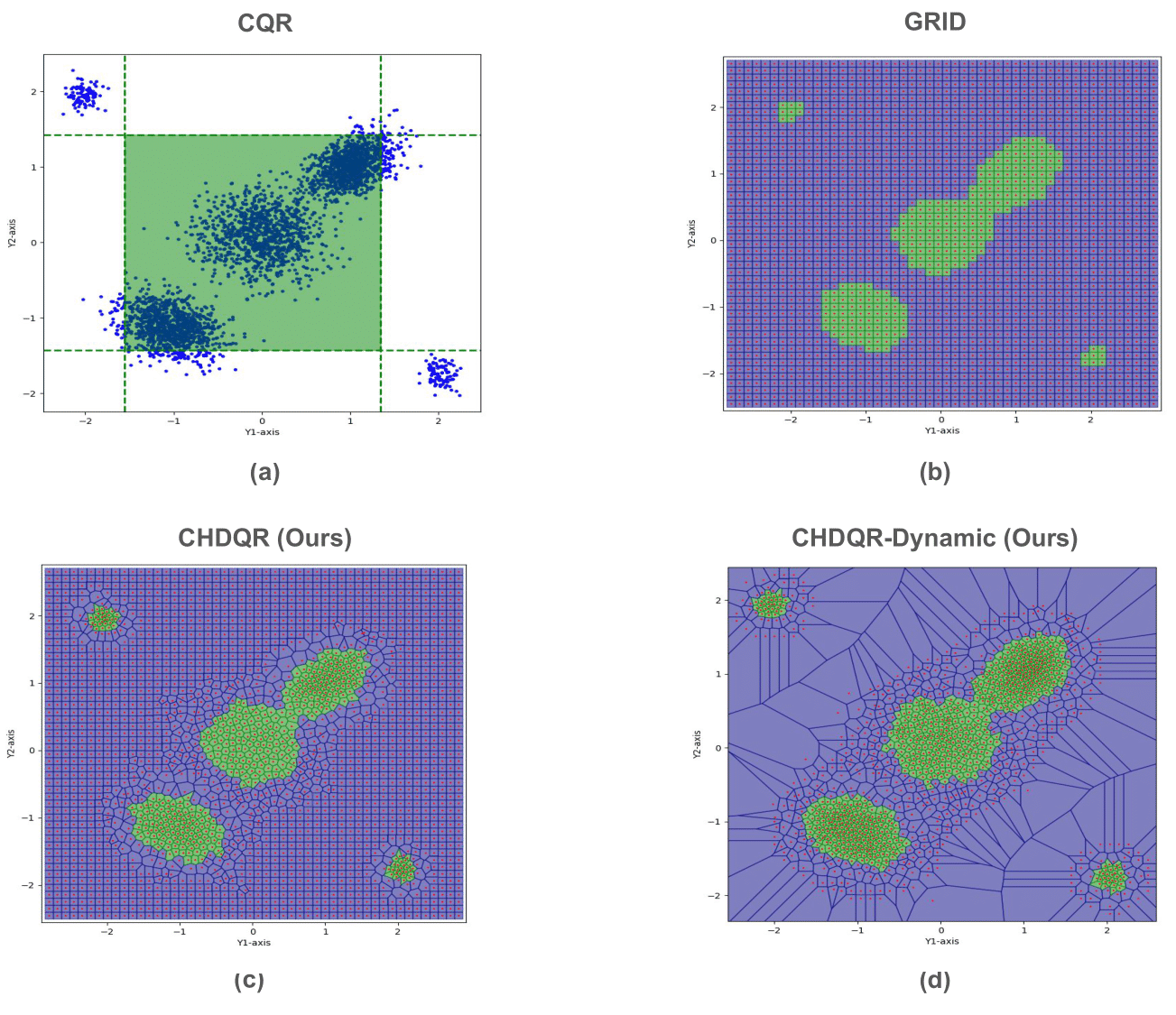}
        \caption{Uncond2d Outlier (6\%) with 90\% Coverage}
        \label{fig:2d_outlier_6}
    \end{subfigure}
    \caption{Visualization of Uncond2d Outlier Datasets with Different Outlier Rates at 90\% Coverage}
    \label{fig:2d_outlier_comparison}
\end{figure}

Notably, CHDQR-Dynamic achieves these results with significantly fewer clusters (ranging from 1301 to 1257), whereas GRID and CQR require 2500 clusters to reach comparable coverage. This efficiency becomes even more evident with added outliers; while CQR experiences an increase in PINAW values, indicating sensitivity to outliers, CHDQR-Dynamic sustains narrow regions and stable performance. GRID, while achieving the target coverage, also shows less efficient region size management due to its high cluster usage.

At the 90\% coverage level, CHDQR-Dynamic maintains the target coverage (0.90) across all configurations, even with increasing outlier percentages, using fewer clusters than GRID and CQR. For example, in the 6\% outlier scenario, CHDQR-Dynamic achieves a lower PINAW of 3.32, outperforming both GRID (3.42) and CHDQR (3.35) while using fewer clusters.

Figures \ref{fig:2d_outlier_0.6} show CHDQR-Dynamic’s adaptability in handling outliers by dynamically assigning clusters and adjusting their positions to better represent these data points, while GRID and CQR appear visually unaffected by outliers. In the 6\% outlier case (Figure \ref{fig:2d_outlier_6}), all models incorporate this part of the distribution, yet CHDQR-Dynamic differentiates it with fewer clusters than CHDQR. GRID learns the distribution with static clusters, while CQR remains visually unaffected by the outliers.

%% Error formulas, needs fix
%%For single-output benchmarks, we employ the SCP, UR, and WUR frameworks. The coverage is assessed by the difference between the desired coverage and the Prediction Interval Coverage Probability (PICP) as shown below where $\mathbb{I}$ is the indicator function.
%%\begin{equation}
%%    \text{PICP} = \frac{1}{T} \sum_{t=1}^{T} \mathbb{I} \left( y_t \in [\underline{\hat{y}}_t , \overline{\hat{y}}_t] \right)
%%\end{equation}
%%and the Prediction Interval Normalized Average Width (PINAW) is used to report the interval width.
%%\begin{equation}
%%    \text{PINAW} = \frac{1}{T} \sum_{t=1}^{T} \frac{\overline{\hat{y}}_t - \underline{\hat{y}}_t}{\max(y) - \min(y)}
%%    = \oint_{\Gamma(y)} dV
%%    = \frac{1}{\|D_{test}\|} \sum_{\|D_{test}\|} \sum_{\Gamma(y)} A(\Gamma(y)_i)
%%\end{equation}

%%As we have multiple targets, we evaluate the Coverage by calculating the miscoverage of the Prediction Region Coverage Probability (PRCP), i.e. simultaneous coverage, as defined below:
%%\begin{equation}
%%    \text{PRCP} = \frac{1}{T} \sum_{t=1}^{T} \mathbb{I} \left( y_t^m \in [\underline{\hat{y}}_t^m , \overline{\hat{y}}_t^m], \forall \: m \in \{1, \dots, M\} \right)
%%\end{equation}
%%We also report the PR, which represents the volume of the hyper-rectangle, using the following expression:
%%\begin{equation}
 %%   \text{PR} = \frac{1}{T} \sum_{t=1}^{T} \prod_{m = 1} ^ M \hat{C}_{\delta_{m}}^{m} (u_t)
%%\end{equation}

% \section{Discussion \& Limitations}
\section{Discussion \& Conclusion} \label{sec_conc}
In this paper, we introduced CHDQR for high-density quantile regression, a method designed to define adaptive prediction regions based on data density. We proposed dynamic prototypes to adjust the number of bins in high-dimensional spaces and introduced a conformalization procedure to determine regions with the highest density for a target quantile rate $1 - \alpha$ with marginal coverage guarantee. Our results demonstrate that CHDQR is scalable to higher dimensions, adaptively defines smaller, more precise regions, and consistently achieves the desired coverage rates. Additionally, we showed that CHDQR effectively manages outliers and diverse data distributions, maintaining robustness across varying levels of data complexity. Compared to previous quantile regression methods using a classifier, CHDQR achieves narrower prediction regions with fewer clusters, demonstrating efficiency both in computational demands and in predictive precision in higher target dimensions. This work opens avenues for further applications in high-dimensional data environments, where adaptive, density-based approaches are increasingly valuable.

% Limitations here

%Introduced a lot of hyperparameters and more complex structures than baseline methods. Voroni calculations require some CPU computation complexity. Explicit computation of area and density for probability calculation may introduce some error. Our current voroni does not work for the more than 3 dimension we also consider some soft area calculation but it will also may introduce more error. Makalede 2 den fazla dimension kullanmadık, daha güzel bir alan hesabı kullanılarak 2 den fazla dimension için güzel bir şekilde denenbilir. For the add-drop prototypes more sophisticated methods may be tried.  

%Limitations:
Our method introduces multiple hyperparameters and a more complex structure compared to baseline approaches, which may impact ease of use and tuning. The Voronoi calculations add to CPU computation complexity, and the explicit area and density calculations required for probability estimation may introduce minor errors. Currently, the Voronoi approach is restricted to three dimensions, limiting its application in higher-dimensional spaces.

%Future Work:
Future improvements could include exploring alternative area calculation methods, such as soft area estimations, to extend applicability to higher dimensions while minimizing error. Additionally, refining the area calculation approach would enable more robust experimentation in spaces with more than two dimensions. For the add-drop prototypes, incorporating more sophisticated selection methods could further enhance model performance and flexibility.

%...
%Conclusion
%In this paper, we introduced CHDQR for high density quantile regression. We suggested dynamic prototypes to adjust the number of bins in high dimensitional space based on density and defined a conformalization procedure to determine the set of regions with the highest density for desired quantile rate $1-\alpha$. We showed our method is scaleable for higher dimensions and able to construct proper regions. CHDQR adaptively defines a smaller region while still achieving desired covarage rates. 

\section*{Declaration of generative AI and AI-assisted technologies in the writing process}
During the preparation of this work, the author(s) used ChatGPT in order to improve language and readability. After using this tool/service, the author(s) reviewed and edited the content as needed and take(s) full responsibility for the content of the publication.

%% The Appendices part is started with the command \appendix;
%% appendix sections are then done as normal sections
\newpage
\appendix
\section{Sythetic Data Details} \label{app:Sythetic}

To make the dataset unconditional, when mapping x  to y  using a neural network, we input a constant value of zero as the feature, allowing the model to generate outputs without dependence on external features. This setup allows for a clearer evaluation of each method’s ability to adapt to multi-dimensional, complex data distributions.

The probability density function (PDF) for {Uncond2d} is defined as:

\[
\text{{Uncond2d}:} [Y_1,Y_2] \sim \frac{1}{3} \mathcal{N} \left(\begin{bmatrix} 0 \\ 0 \end{bmatrix}, \begin{bmatrix} 1 & 0 \\ 0 & 1 \end{bmatrix}\right) + \frac{1}{3} \mathcal{N} \left(\begin{bmatrix} 3 \\ 3 \end{bmatrix}, \begin{bmatrix} 0.5 & 0.2 \\ 0.2 & 0.5 \end{bmatrix}\right) + \frac{1}{3} \mathcal{N} \left(\begin{bmatrix} -3 \\ -4 \end{bmatrix}, \begin{bmatrix} 0.7 & -0.2 \\ -0.2 & 0.5 \end{bmatrix}\right)
\]

The probability density function (PDF) for {Uncond1d} is defined as: 
\[
\text{{Uncond1d}:} [Y_1] \sim \frac{1}{2} \mathcal{N}(0.75, 0.05) + \frac{1}{2} \mathcal{N}(-0.75, 0.05)
\]
\input{tables/table_2d_outlier}

% \section*{List of Variables \color{red} (Will be deleted!)}

% \begin{table}[htbp]
% \centering
% \renewcommand{\arraystretch}{0.9} % Adjust this value to make the rows tighter
% \setlength{\tabcolsep}{4pt} % Adjust column separation (optional)
% \begin{tabular}{ll}
% \textbf{Variable} & \textbf{Description} \\ \hline
% $y$  & True value of y. \\
% $\hat{y}$  & Prediction of y. \\
% $q(y, \tau)$  & assigned probability vector of y, quantization vector. \\ 
% $c_i$ & Protototypes. \\
% $R_i$ & Region of protototype $c_i$. \\
% $A_i$ & Size of Region $R_i$. \\
% $p_i$ & Probability density of $P(y \in R_i) = p_i * A_i$. \\
% $\hat{P}$ & Predicted probability vector. \\
% $f(x;\theta)$ & Log likelihood estimator, neural network. \\
% $\mathcal{L}_{CE}$ & Cross-entropy Loss \\
% $\mathcal{L}_{q}$ & MinDist Loss \\
% $\mathcal{L}_{rep}$ & Repulsion Loss \\
% \end{tabular}
% \end{table}

\newpage
\bibliographystyle{elsarticle-num} 
\bibliography{refs}

% \end{thebibliography}
\end{document}

%% file: algorithms/MainLoop.tex
\begin{algorithm}[ht]
\caption{Main Training Loop for CHDQR}
\label{Algo: Main_loop}
\begin{algorithmic}[1]
\State \textbf{Input:} Log-density estimator $f(x;\theta)$, Prototypes $\mathcal{C}$, Training Data $\mathcal{D}_{tr}$
\State \textbf{Hyperparameters:} Epochs $T$, Add Period $T_{add}$, Remove Proto Period $T_{del}$, softlabel temperature $\tau$, Repulsion margin $\delta_{rep}$, Deletion threshold $\delta_{del}$, Addition threshold $\delta_{add}$

\For{$t = 1$ to $T$}
    \For{Every $T_{add}$ epochs} 
        \State \Call{AddProto}{$f(x;\theta), \mathcal{C}, \mathcal{D}_{tr}$}
        \Comment{See Algorithm 2}
    \EndFor
    \For{Every $T_{del}$ epochs} 
        \State \Call{RemoveProto}{$f(x;\theta), \mathcal{C}, \mathcal{D}_{tr}$}
        \Comment{See Algorithm 2}
    \EndFor

    \For{each mini-batch $(\mathbf{x}, \mathbf{y}) \in \mathcal{D}_{tr}$}
        \State $q(y,\tau) \gets$ Calculate soft label vector \Comment{Eq. \eqref{eq:def_labeling}} 
        
        \State $\mathbf{A}_i \gets$  Calculate Areas using \Call{Voronoi}{$\mathcal{C}$} % \Comment{Eq. \eqref{eq:approximators}}
        
        \State $\log p(x) \gets$ Forward pass through $f(x;\theta)$ \Comment{Eq. \eqref{eq:approximators}}
        
        \State $\hat{P}[y \in R_i \mid x] \gets$ Calculate prediction vector\Comment{Eq. \eqref{eq:softmax_pred}}

        \State $\mathcal{L} \gets$ Calculate loss \Comment{Eq. \eqref{eq:total_loss}}

        \State $\theta^*, \mathcal{C^*} \gets$ Update parameters w.r.t. $\mathcal{L}$

    \EndFor
\EndFor
\end{algorithmic}
\end{algorithm}

%% file: algorithms/AddRemoveProto.tex
\begin{algorithm}[ht] 
\caption{Add/Remove Prototypes during Training}
\label{Algo: Add_Remove_Proto}
\begin{algorithmic}[1]
\Function{RemoveProto}{$f(x;\theta), \mathcal{C}, \mathcal{D}_{tr}$}
    \For{each Prototype $c_i \in \mathcal{C}$ }
    \State Compute $U(c_i)$
    \Comment{Eq. \eqref{eq:usage}}
    
    \If{$U(c_i) \leq \delta_{\text{del}}$ } 
    \State Update $\mathcal{C} \longleftarrow \mathcal{C} - \{ c_{i} \}  $
    \Comment{Eq. \eqref{eq:delete_proto}}
    
    \State Update $W \longleftarrow W - \{ w_i \}, b \longleftarrow b - \{ b_i \}$
    \Comment{Eq. \eqref{eq:delete_weight}}
    \EndIf
    \EndFor
\EndFunction

\Function{AddProto}{$f(x;\theta), \mathcal{C}, \mathcal{D}_{tr}$}
    \For{each Prototype $c_i \in \mathcal{C}$ }\State Compute $U(c_i)$
    \Comment{Eq. \eqref{eq:usage}}
    
    \If{$U(c_i) \geq \delta_{\text{add}}$ } 
    \State Sample Noise $\epsilon \sim \mathcal{N}(0, \sigma^2)$
    \State New prototype $c_{K+1} \longleftarrow c_{i} + \epsilon$
    
    \State Update $\mathcal{C} \longleftarrow \{ \mathcal{C}, c_{K+1} \}$
    \Comment{Eq. \eqref{eq:add_proto}}
    
    \State Update $W \longleftarrow \text{concat}(W, \{ w_{K+1} \}), b \longleftarrow \text{concat}(b, \{ b_{K+1} \})$
    \Comment{Eq. \eqref{eq:add_weight}}
    
    \EndIf
    \EndFor
\EndFunction

\end{algorithmic}
\end{algorithm}

%% file: algorithms/Conformalize.tex
% \begin{algorithm}[t] 
% \caption{Conformalized Prediction Procedure}
% \label{Algo: Conformal_Prediction}
% \begin{algorithmic}[1]

% \Function{ConformalPrediction}{$\mathcal{D}_{cal}, \mathcal{D}_{test}, \alpha, f(x;\theta), \mathcal{C}$}
%     \State \textbf{Calibration:}
%     \State Compute nonconformity scores $S(y, x)$ for all $(x, y) \in \mathcal{D}_{cal}$
%     \State Calculate the quantile threshold $\hat{q}_{\alpha}$ using $S(y, x)$
    
%     \State \textbf{Prediction:}
%     \For{each test sample $(x_{test}, y_{test}) \in \mathcal{D}_{test}$}
%         \State Compute $S(y_{test}, x_{test})$
%         \State $\Gamma(x_{test}) = \{y \in \mathbb{R}^d : S(y, x_{test}) \leq \hat{q}_{\alpha}\}$
%     \EndFor

%     % \State Calculate coverage and PINAW
% \EndFunction

% \end{algorithmic}
% \end{algorithm}

\begin{algorithm}[t] 
\caption{Conformalized High Density Quantile Regression}
\label{Algo: Conformal_Prediction}
\begin{algorithmic}[1]

\Function{ConformalPrediction}{$\mathcal{D}_{cal}, x_{test}, \alpha, f(x;\theta), \mathcal{C}$}
    \State \textbf{Calibration:}
    \For{each $(x_{cal}, y_{cal}) \in \mathcal{D}_{cal}$}
        \State $\log p(x_{cal}) \gets$ Forward pass through $f(x_{cal};\theta)$ \Comment{Eq. \eqref{eq:approximators}}
        \State $\pi(x_{cal}) \gets$ Sort prototypes by densities; $\text{argsort}(p(x_{cal}))$
        \State $\rho(x_{cal}) \gets$ Calculate cumulative probabilities  \Comment{Eq. \eqref{eq:cumulative_probs}}
        \State $S(y_{cal}, x_{cal}) \gets$ Compute nonconformity scores \Comment{Eq. \eqref{eq:nonconf_score}} 
    \EndFor
    \State $\hat{q}_{\alpha} \gets$ Calculate $1-\alpha$'th quantile \Comment{Eq. \eqref{eq:quantile_qhat}}
    \\
    \State \textbf{Prediction:}
    \For{New sample $x_{test}$}
        \State $\log p(x_{test}) \gets$ Forward pass through $f(x_{test};\theta)$ \Comment{Eq. \eqref{eq:approximators}}
        \State $\pi(x_{test}) \gets$ Sort prototypes by densities; $\text{argsort}(p(x_{test}))$
        \State $\rho(x_{test}) \gets$ Calculate cumulative probabilities  \Comment{Eq. \eqref{eq:cumulative_probs}}
        \State $\Gamma(x_{test}) \gets$ Build prediction set $\left\{ R_{\pi_i} \mid \rho(x_{test}) \leq \hat{q}_{\alpha} \right\}$ \Comment{Eq. \eqref{eq:conformal_prediction}}
    \EndFor

\EndFunction

\end{algorithmic}
\end{algorithm}

%% file: tables/table_1d_90.tex
\begin{table}[ht]     \caption{Results of CQR \cite{romano2019conformalized}, GRID \cite{guha2024conformal}, CHDQR, and CHDQR-Dynamic on 1D Datasets for 90\% Target Coverage}
    \small
    \centering
    \renewcommand{\arraystretch}{1.2}
    \scalebox{0.8}{
    \hspace*{-1cm}
    \begin{tabular}{llccccccc}
        \toprule
        \textbf{Method} & \textbf{Measure} & \textbf{Bike} & \textbf{Concrete} & \textbf{MEPS19} & \textbf{MEPS20} & \textbf{MEPS21} & \textbf{Wine} & \textbf{Uncond1d} \\
        \midrule
        \multirow{2}{*}{CQR} 
        & Coverage & $0.90_{(0.01)}$ & $0.88_{( 0.02)}$ & $0.90_{(0.01)}$ & $0.90_{0.01}$ & $0.90_{(0.01 )}$ & $0.90_{(0.01)}$ & $0.90_{(0.01)}$ \\
        & PINAW & $2.99_{(0.18)}$ & $2.98_{(0.16)}$ & $0.71_{(0.07)}$ & $0.74_{(0.05)}$ & $0.72_{(0.07)}$ & $2.89_{(0.22)}$ & $2.42_{(0.22)}$ \\
        \midrule
        \multirow{3}{*}{GRID} 
        & Coverage & $0.91_{(0.01)}$ & $0.92_{(0.02)}$ & $0.90_{(0.01)}$ & $0.90_{(0.01)}$ & $0.90_{(0.01)}$ & $0.90_{(0.02)}$ & $0.92_{(0.02)}$ \\
        & PINAW & $2.23_{(0.07)}$ & $1.23_{(0.09)}$ & $0.84_{(0.03)}$ & $0.85_{(0.04)}$ & $0.82_{(0.06)}$ & $0.79_{(0.06)}$ & $0.50_{(0.02)}$ \\
        & Cluster Number & 50 & 50 & 50 & 50 & 50 & 50 & 50 \\
        \midrule
        \multirow{3}{*}{CHDQR} 
        & Coverage & $0.91_{(0.01)}$ & $0.93_{(0.02)}$ & $0.90_{(0.01)}$ & $0.90_{(0.01)}$ & $0.90_{(0.01)}$ & $0.90_{(0.01)}$ & $0.91_{(0.01)}$ \\
        & PINAW & $2.22_{(0.08)}$ & $1.23_{(0.09)}$ & $0.84_{(0.04)}$ & $0.84_{(0.06)}$ & $0.81_{(0.08)}$ & $0.56_{(0.03)}$ & $0.46_{(0.01)}$ \\
        & Cluster Number & 50 & 50 & 50 & 50 & 50 & 50 & 50 \\
        \midrule
        \multirow{3}{*}{CHDQR-Dynamic} 
        & Coverage & $0.90_{(0.01)}$ & $0.91_{(0.03)}$ & $0.90_{(0.01)}$ & $0.90_{(0.01)}$ & $0.90_{(0.01)}$ & $0.90_{(0.02)}$ & $0.92_{(0.01)}$ \\
        & PINAW & $2.01_{(0.09)}$ & $1.18_{(0.10)}$ & $0.82_{(0.05)}$ & $0.90_{(0.10)}$ & $0.75_{(0.10)}$ & $0.41_{(0.07)}$ & $0.48_{(0.01)}$ \\
        & Cluster Number & 142 & 93 & 33 & 35 & 33 & 66 & 85 \\
        \bottomrule
    \end{tabular}
    }
    \label{tab:1d-comparison-0.9}
\end{table}

%% file: tables/table_1d_50.tex
\begin{table}[ht]     \caption{Results of CQR \cite{romano2019conformalized}, GRID \cite{guha2024conformal}, CHDQR, and CHDQR-Dynamic on 1D Datasets for 50\% Target Coverage}
    \small
    \centering
    \renewcommand{\arraystretch}{1.2}
    \scalebox{0.8}{
    \hspace*{-1cm}
    \begin{tabular}{llccccccc}
        \toprule
        \textbf{Method} & \textbf{Measure} & \textbf{Bike} & \textbf{Concrete} & \textbf{MEPS19} & \textbf{MEPS20} & \textbf{MEPS21} & \textbf{Wine} & \textbf{Uncond1d} \\
        \midrule
        \multirow{2}{*}{CQR} 
        & Coverage & $0.50_{(0.02)}$ & $0.50_{(0.07)}$ & $0.50_{(0.02)}$ & $0.51_{(0.02)}$ & $0.50_{(0.02)}$ & $0.49_{(0.03)}$ & $0.51_{(0.01)}$ \\
        & PINAW & $1.22_{(0.05)}$ & $1.31_{(0.16)}$ & $0.22_{(0.03)}$ & $0.23_{(0.02)}$ & $0.21_{(0.01)}$ & $1.48_{(0.29)}$ & $1.75_{(0.23)}$ \\
        \midrule
        \multirow{3}{*}{GRID} 
        & Coverage & $0.50_{(0.02)}$ & $0.55_{(0.07)}$ & $0.51_{(0.02)}$ & $0.52_{(0.02)}$ & $0.51_{(0.01)}$ & $0.51_{(0.03)}$ & $0.60_{(0.01)}$ \\
        & PINAW & $0.60_{(0.04)}$ & $0.42_{(0.07)}$ & $0.50_{(0.02)}$ & $0.50_{(0.02)}$ & $0.48_{(0.04)}$ & $0.24_{(0.03)}$ & $0.24_{(0.00)}$ \\
        & Cluster Number & 50 & 50 & 50 & 50 & 50 & 50 & 50 \\
        \midrule
        \multirow{3}{*}{CHDQR} 
        & Coverage & $0.50_{(0.03)}$ & $0.55_{(0.06)}$ & $0.50_{(0.01)}$ & $0.51_{(0.02)}$ & $0.50_{(0.02)}$ & $0.50_{(0.03)}$ & $0.52_{(0.02)}$ \\
        & PINAW & $0.60_{(0.04)}$ & $0.43_{(0.06)}$ & $0.31_{(0.02)}$ & $0.30_{(0.01)}$ & $0.30_{(0.02)}$ & $0.15_{(0.01)}$ & $0.19_{(0.00)}$ \\
        & Cluster Number & 50 & 50 & 50 & 50 & 50 & 50 & 50 \\
        \midrule
        \multirow{3}{*}{CHDQR-Dynamic} 
        & Coverage & $0.50_{(0.02)}$ & $0.55_{(0.06)}$ & $0.50_{(0.02)}$ & $0.50_{(0.02)}$ & $0.50_{(0.01)}$ & $0.49_{(0.04)}$ & $0.52_{(0.03)}$ \\
        & PINAW & $0.55_{(0.03)}$ & $0.40_{(0.06)}$ & $0.17_{(0.05)}$ & $0.24_{(0.04)}$ & $0.17_{(0.04)}$ & $0.08_{(0.03)}$ & $0.16_{(0.02)}$ \\
        & Cluster Number & 142 & 93 & 33 & 35 & 33 & 66 & 85 \\
        \bottomrule
    \end{tabular}
    }
    \label{tab:1d-comparison-0.5}
\end{table}

%% file: tables/table_1d_10.tex
\begin{table}[ht]    \caption{Results of CQR \cite{romano2019conformalized}, GRID \cite{guha2024conformal}, CHDQR, and CHDQR-Dynamic on 1D Datasets for 10\% Target Coverage}
    \small
    \centering
    \renewcommand{\arraystretch}{1.2}
    \scalebox{0.8}{
    \hspace*{-1cm}
    \begin{tabular}{llccccccc}
        \toprule
        \textbf{Method} & \textbf{Measure} & \textbf{Bike} & \textbf{Concrete} & \textbf{MEPS19} & \textbf{MEPS20} & \textbf{MEPS21} & \textbf{Wine} & \textbf{Uncond1d} \\
        \midrule
        \multirow{2}{*}{CQR} 
        & Coverage & $0.10_{(0.01)}$ & $0.12_{(0.06)}$ & $0.13_{(0.01)}$ & $0.13_{(0.02)}$ & $0.13_{(0.01)}$ & $0.11_{(0.01)}$ & $0.10_{(0.01)}$ \\
        & PINAW & $0.21_{(0.02)}$ & $0.24_{(0.06)}$ & $0.04_{(0.01)}$ & $0.04_{(0.01)}$ & $0.03_{(0.01)}$ & $0.14_{(0.03)}$ & $1.22_{(0.20)}$ \\
        \midrule
        \multirow{3}{*}{GRID} 
        & Coverage & $0.10_{(0.01)}$ & $0.15_{(0.04)}$ & $0.10_{(0.01)}$ & $0.10_{(0.01)}$ & $0.10_{(0.01)}$ & $0.10_{(0.02)}$ & $0.20_{(0.01)}$ \\
        & PINAW & $0.12_{(0.01)}$ & $0.09_{(0.02)}$ & $0.13_{(0.01)}$ & $0.13_{(0.01)}$ & $0.13_{(0.01)}$ & $0.05_{(0.01)}$ & $0.06_{(0.00)}$ \\
        & Cluster Number & 50 & 50 & 50 & 50 & 50 & 50 & 50 \\
        \midrule
        \multirow{3}{*}{CHDQR} 
        & Coverage & $0.10_{(0.01)}$ & $0.12_{(0.02)}$ & $0.10_{(0.01)}$ & $0.10_{(0.01)}$ & $0.11_{(0.01)}$ & $0.10_{(0.02)}$ & $0.12_{(0.02)}$ \\
        & PINAW & $0.12_{(0.01)}$ & $0.08_{(0.04)}$ & $0.06_{(0.00)}$ & $0.07_{(0.00)}$ & $0.07_{(0.01)}$ & $0.04_{(0.00)}$ & $0.04_{(0.01)}$ \\
        & Cluster Number & 50 & 50 & 50 & 50 & 50 & 50 & 50 \\
        \midrule
        \multirow{3}{*}{CHDQR-Dynamic} 
        & Coverage & $0.10_{(0.01)}$ & $0.13_{(0.05)}$ & $0.10_{(0.01)}$ & $0.10_{(0.01)}$ & $0.10_{(0.01)}$ & $0.10_{(0.02)}$ & $0.11_{(0.01)}$ \\
        & PINAW & $0.07_{(0.01)}$ & $0.06_{(0.01)}$ & $0.01_{(0.01)}$ & $0.03_{(0.01)}$ & $0.01_{(0.01)}$ & $0.01_{(0.01)}$ & $0.02_{(0.01)}$ \\
        & Cluster Number & 142 & 93 & 33 & 35 & 33 & 66 & 85 \\
        \bottomrule
    \end{tabular}
    }
    \label{tab:1d-comparison-0.1}
\end{table}

%% file: tables/table_2d.tex
\begin{table}[h]
    \caption{Comparison of Coverage, PINAW, and Cluster Numbers for Different Datasets and Methods}
    \centering
    \small
    \renewcommand{\arraystretch}{1.2}
    \scalebox{0.85}{
    \begin{tabular}{llcccc}
        \toprule
        \textbf{Dataset} & \textbf{Measure} & \textbf{CQR} & \textbf{GRID} & \textbf{CHDQR} & \textbf{CHDQR-Dynamic} \\
        \midrule
        
        \multirow{3}{*}{\textbf{{Uncond2d} (90\%)}} 
        & Coverage & $0.82_{(0.01)}$ & $0.90_{(0.01)}$ & $0.90_{(0.01)}$ & $0.90_{(0.01)}$ \\
        & PINAW & $8.45_{(0.01)}$ & $3.72_{(0.08)}$ & $3.70_{(0.09)}$ & $3.69_{(0.09)}$ \\
        & Cluster Number & 2500 & 2500 & 2500 & 1301 \\
        \midrule
        \multirow{3}{*}{\textbf{{Uncond2d} (50\%)}} 
        & Coverage & $0.37_{(0.02)}$ & $0.50_{(0.02)}$ & $0.50_{(0.02)}$ & $0.50_{(0.02)}$ \\
        & PINAW & $3.80_{(0.16)}$ & $1.06_{(0.04)}$ & $1.07_{(0.03)}$ & $1.07_{(0.04)}$ \\
        & Cluster Number & 2500 & 2500 & 2500 & 1301 \\
        \midrule
        \multirow{3}{*}{\textbf{{Uncond2d} (10\%)}} 
        & Coverage & $0.03_{(0.00)}$ & $0.11_{(0.01)}$ & $0.10_{(0.01)}$ & $0.10_{(0.01)}$ \\
        & PINAW & $0.08_{(0.01)}$ & $0.14_{(0.01)}$ & $0.14_{(0.01)}$ & $0.14_{(0.01)}$ \\
        & Cluster Number & 2500 & 2500 & 2500 & 1301 \\
        \midrule
        \multirow{3}{*}{\textbf{Energy Efficiency (90\%)}} 
        & Coverage & $0.85_{(0.04)}$ & $0.92_{(0.04)}$ & $0.91_{(0.05)}$ & $0.88_{(0.07)}$ \\
        & PINAW & $7.34_{(0.98)}$ & $0.24_{(0.05)}$ & $0.20_{(0.03)}$ & $0.20_{(0.09)}$ \\
        & Cluster Number & 2500 & 2500 & 2500 & 1319 \\
        \midrule
        \multirow{3}{*}{\textbf{Energy Efficiency (50\%)}} 
        & Coverage & $0.39_{(0.08)}$ & $0.51_{(0.08)}$ & $0.53_{(0.06)}$ & $0.48_{(0.21)}$ \\
        & PINAW & $2.13_{(0.44)}$ & $0.03_{(0.01)}$ & $0.03_{(0.01)}$ & $0.03_{(0.02)}$ \\
        & Cluster Number & 2500 & 2500 & 2500 & 1319 \\
        \midrule
        \multirow{3}{*}{\textbf{Energy Efficiency (10\%)}} 
        & Coverage & $0.07_{(0.04)}$ & $0.12_{(0.05)}$ & $0.11_{(0.06)}$ & $0.11_{(0.03)}$ \\
        & PINAW & $0.16_{(0.09)}$ & $0.01_{(0.00)}$ & $0.01_{(0.00)}$ & $0.01_{(0.00)}$ \\
        & Cluster Number & 2500 & 2500 & 2500 & 1319 \\
        \bottomrule
    \end{tabular}
    }

    \label{tab:combined-comparison}
\end{table}

%% file: tables/table_2d_outlier_90.tex
\begin{table}[ht]     \caption{Comparison of Coverage, PINAW, and Cluster Numbers for {Uncond2d} Dataset with and without outliers across different methods and 90\% target coverage}
    \centering
    \small
    \renewcommand{\arraystretch}{1.2}
    \scalebox{0.85}{
    \begin{tabular}{llcccc}
        \toprule
        \textbf{Dataset} & \textbf{Measure} & \textbf{CQR} & \textbf{GRID} & \textbf{CHDQR } & \textbf{CHDQR-Dynamic } \\
        \midrule
        \multirow{3}{*}{\textbf{{Uncond2d} }} 
        & Coverage  & $0.82_{(0.01)}$ & $0.90_{(0.01)}$ & $0.90_{(0.01)}$ & $0.90_{(0.01)}$ \\
        & PINAW  & $8.45_{(0.01)}$ & $3.72_{(0.08)}$ & $3.70_{(0.09)}$ & $3.69_{(0.09)}$ \\
        & Cluster Number & 2500 & 2500 & 2500 & 1301 \\
        \midrule
        \multirow{3}{*}{\textbf{\begin{tabular}[c]{@{}l@{}}Uncond2d \\ (0.6\% Outlier) \\ \end{tabular}}} 
        & Coverage  & $0.82_{(0.00)}$ & $0.90_{(0.01)}$ & $0.90_{(0.01)}$ & $0.90_{(0.01)}$ \\
        & PINAW  & $8.53_{(0.24)}$ & $3.73_{(0.10)}$ & $3.72_{(0.10)}$ & $3.69_{(0.08)}$ \\
        & Cluster Number & 2500 & 2500 & 2500 & 1114 \\
        \midrule
        \multirow{3}{*}{\textbf{\begin{tabular}[c]{@{}l@{}}Uncond2d \\ (6\% Outlier) \\ \end{tabular}}} 
        & Coverage  & $0.87_{(0.01)}$ & $0.90_{(0.01)}$ & $0.90_{(0.01)}$ & $0.90_{(0.01)}$ \\
        & PINAW  & $8.65_{(0.40)}$ & $3.42_{(0.08)}$ & $3.35_{(0.06)}$ & $3.32_{(0.07)}$ \\
        & Cluster Number & 2500 & 2500 & 2500 & 1257 \\
        \bottomrule
    \end{tabular}
    }
    \label{tab:outlier-comparison}
\end{table}

%% file: tables/table_2d_outlier.tex
\begin{table}[H]
    \caption{Comparison of Coverage, PINAW, and Cluster Numbers for {Uncond2d} Dataset with and without Outliers across Different Methods and Coverage Levels}

    \centering
    \small
    \renewcommand{\arraystretch}{1.2}
    \scalebox{0.85}{
    \begin{tabular}{llcccc}
        \toprule
        \textbf{Dataset} & \textbf{Measure} & \textbf{CQR} & \textbf{GRID} & \textbf{CHDQR } & \textbf{CHDQR-Dynamic } \\
        \midrule
        \multirow{3}{*}{\textbf{{Uncond2d} (90\%)}} 
        & Coverage  & $0.82_{(0.01)}$ & $0.90_{(0.01)}$ & $0.90_{(0.01)}$ & $0.90_{(0.01)}$ \\
        & PINAW  & $8.45_{(0.01)}$ & $3.72_{(0.08)}$ & $3.70_{(0.09)}$ & $3.69_{(0.09)}$ \\
        & Cluster Number & 2500 & 2500 & 2500 & 1301 \\
        \midrule
        \multirow{3}{*}{\textbf{{Uncond2d} Outlier (0.6\%) (90\%)}}
        & Coverage  & $0.82_{(0.00)}$ & $0.90_{(0.01)}$ & $0.90_{(0.01)}$ & $0.90_{(0.01)}$ \\
        & PINAW  & $8.53_{(0.24)}$ & $3.73_{(0.10)}$ & $3.72_{(0.10)}$ & $3.69_{(0.08)}$ \\
        & Cluster Number & 2500 & 2500 & 2500 & 1257 \\
        \midrule
        \multirow{3}{*}{\textbf{{Uncond2d} Outlier (6\%) (90\%)}} 
        & Coverage  & $0.87_{(0.01)}$ & $0.90_{(0.01)}$ & $0.90_{(0.01)}$ & $0.90_{(0.01)}$ \\
        & PINAW  & $8.65_{(0.40)}$ & $3.42_{(0.08)}$ & $3.35_{(0.06)}$ & $3.32_{(0.07)}$ \\
        & Cluster Number & 2500 & 2500 & 2500 & 1257 \\
        \midrule
        \multirow{3}{*}{\textbf{{Uncond2d} (50\%)}} 
        & Coverage  & $0.37_{(0.02)}$ & $0.50_{(0.02)}$ & $0.50_{(0.02)}$ & $0.50_{(0.02)}$ \\
        & PINAW  & $3.80_{(0.16)}$ & $1.06_{(0.04)}$ & $1.07_{(0.03)}$ & $1.07_{(0.04)}$ \\
        & Cluster Number & 2500 & 2500 & 2500 & 1301 \\
        \midrule
        \multirow{3}{*}{\textbf{{Uncond2d} Outlier (0.6\%) (50\%)}} 
        & Coverage  & $0.37_{(0.02)}$ & $0.50_{(0.01)}$ & $0.50_{(0.02)}$ & $0.50_{(0.01)}$ \\
        & PINAW  & $3.74_{(0.08)}$ & $1.05_{(0.02)}$ & $1.05_{(0.03)}$ & $1.04_{(0.03)}$ \\
        & Cluster Number & 2500 & 2500 & 2500 & 1257 \\
        \midrule
        \multirow{3}{*}{\textbf{{Uncond2d} Outlier (6\%) (50\%)}} 
        & Coverage  & $0.37_{(0.02)}$ & $0.51_{(0.01)}$ & $0.51_{(0.02)}$ & $0.51_{(0.02)}$ \\
        & PINAW  & $3.37_{(0.11)}$ & $0.97_{(0.03)}$ & $0.96_{(0.03)}$ & $0.96_{(0.03)}$ \\
        & Cluster Number & 2500 & 2500 & 2500 & 1257 \\
        \midrule
        \multirow{3}{*}{\textbf{{Uncond2d} (10\%)}} 
        & Coverage  & $0.03_{(0.00)}$ & $0.11_{(0.01)}$ & $0.10_{(0.01)}$ & $0.10_{(0.01)}$ \\
        & PINAW  & $0.08_{(0.01)}$ & $0.14_{(0.01)}$ & $0.14_{(0.01)}$ & $0.14_{(0.01)}$ \\
        & Cluster Number & 2500 & 2500 & 2500 & 1301 \\
        \midrule
        \multirow{3}{*}{\textbf{{Uncond2d} Outlier (0.6\%) (10\%)}} 
        & Coverage  & $0.03_{(0.00)}$ & $0.11_{(0.01)}$ & $0.11_{(0.01)}$ & $0.10_{(0.00)}$ \\
        & PINAW  & $0.08_{(0.01)}$ & $0.14_{(0.01)}$ & $0.14_{(0.01)}$ & $0.14_{(0.01)}$ \\
        & Cluster Number & 2500 & 2500 & 2500 & 1257 \\
        \midrule
        \multirow{3}{*}{\textbf{{Uncond2d} Outlier (6\%) (10\%)}} 
        & Coverage  & $0.03_{(0.00)}$ & $0.10_{(0.01)}$ & $0.10_{(0.01)}$ & $0.10_{(0.01)}$ \\
        & PINAW  & $0.07_{(0.01)}$ & $0.12_{(0.01)}$ & $0.12_{(0.01)}$ & $0.12_{(0.01)}$ \\
        & Cluster Number & 2500 & 2500 & 2500 & 1257 \\
        \bottomrule
    \end{tabular}
    }
    \label{tab:outlier-comparison}
\end{table}